\documentclass[utf8]{FrontiersinHarvard} % for articles in journals using the Harvard Referencing Style (Author-Date), for Frontiers Reference Styles by Journal: https://zendesk.frontiersin.org/hc/en-us/articles/360017860337-Frontiers-Reference-Styles-by-Journal
\usepackage{url,hyperref,lineno,microtype,subcaption}
\usepackage[onehalfspacing]{setspace}
\usepackage{multirow}
\usepackage{tablefootnote}

\def\keyFont{\fontsize{8}{11}\helveticabold }
\def\firstAuthorLast{\v{S}tajner {et~al.}} %use et al only if is more than 1 author
\def\Authors{Sanja \v{S}tajner$^{1}$, Daniel Ferr\'{e}s$^{2}$, Matthew Shardlow$^{3,*}$, Kai North$^{4}$, Marcos Zampieri$^{4}$ and Horacio Saggion$^{2}$}
\def\Address{$^{1}$Karlsruhe, Germany \\
$^{2}$LaSTUS Lab, TALN Group, Universitat Pompeu Fabra, Barcelona, Spain\\
$^{3}$Department of Computing and Mathematics,
Manchester Metropolitan University, Manchester, United Kingdom\\
$^{4}$School of Computing, George Mason University, Fairfax, VA, USA}
\def\corrAuthor{Matthew Shardlow, Department of  Computing and Mathematics,
Manchester Metropolitan University, E150, John Dalton East, Chester Street, M1 5GD, United Kingdom}
\def\corrEmail{M.Shardlow@mmu.ac.uk}

\begin{document}
\onecolumn
\firstpage{1}

\title[Lexical Simplification Benchmarks for English, Portuguese, and Spanish]{Lexical Simplification Benchmarks for English, Portuguese, and Spanish} 

\author[\firstAuthorLast ]{\Authors} %This field will be automatically populated
\address{} %This field will be automatically populated
\correspondance{} %This field will be automatically populated

\extraAuth{}% If there are more than 1 corresponding author, comment this line and uncomment the next one.
%\extraAuth{corresponding Author2 \\ Laboratory X2, Institute X2, Department X2, Organization X2, Street X2, City X2 , State XX2 (only USA, Canada and Australia), Zip Code2, X2 Country X2, email2@uni2.edu}

%TC:ignore
\maketitle

\begin{abstract}

Even in highly-developed countries, as many as 15-30\% of the population can only understand texts written using a basic vocabulary. Their understanding of everyday texts is limited, which prevents them from taking an active role in society and making informed decisions regarding healthcare, legal representation, or democratic choice. 

Lexical simplification is a natural language processing task that aims to make text understandable to everyone by replacing complex vocabulary and expressions with simpler ones, while preserving the original meaning. It has attracted considerable attention in the last 20 years, and fully automatic lexical simplification systems have been proposed for various languages. The main obstacle for the progress of the field is the absence of high-quality datasets for building and evaluating lexical simplification systems. 

In this study, we present a new benchmark dataset for lexical simplification in English, Spanish, and (Brazilian) Portuguese, and provide details about data selection and annotation procedures, to enable compilation of comparable datasets in other languages and domains. As the first multilingual lexical simplification dataset, where instances in all three languages were selected and annotated using comparable procedures, this is the first dataset that offers a direct comparison of lexical simplification systems for three languages. To showcase the usability of the dataset, we adapt two state-of-the-art lexical simplification systems with differing architectures (neural vs.\ non-neural) to all three languages (English, Spanish, and Brazilian Portuguese) and evaluate their performances on our new dataset. For a fairer comparison, we use several evaluation measures which capture varied aspects of the systems' efficacy, and discuss their strengths and weaknesses.

We find that a state-of-the-art neural lexical simplification system outperforms a state-of-the-art non-neural lexical simplification system in all three languages, according to all evaluation measures. More importantly, we find that the state-of-the-art neural lexical simplification systems perform significantly better for English than for Spanish and Portuguese, thus posing a question if such an architecture can be used for successful lexical simplification in other languages, especially the low-resourced ones.

\tiny
 \keyFont{ \section{Keywords:} natural language processing, lexical simplification, benchmark datasets, evaluation methodologies, low-resource tasks, artificial intelligence for social good} 
\end{abstract}

\section{Introduction}

According to the adult literacy report conducted in 24 highly-developed countries \citep{OECD-13}, 16.7\% of a population, on average, cannot understand texts that go beyond a basic vocabulary. This percentage is even higher for some countries, e.g.\ 21.7\% for the U.S., and 28.3\% for Spain \textcolor{black}{\citep{OECD-13}}. People who do not correctly understand written information cannot make informed decisions regarding critical processes such as healthcare choices, legal representation, education, or democratic rights. This prevents them from taking an active role in society. 

The disparity between the typical level of vocabulary in written communications and the audiences they were intended for was already evident almost a century ago \citep{ogden37}. Since then, various campaigns advocated for producing easy-to-read texts that would be understood by more people. 
Similarly, in scenarios where critical information needs to be unambiguously communicated, such as technical manual writing or in disaster relief efforts, the use of controlled languages has been suggested \citep{temnikova2015emterms}. 

Although the efforts to promote the use of plain English have been effective in communicating the need for easily understandable text, there has been little consensus on what standards should be adopted and a low uptake of existing standards.
Basic English \citep{ogden37} was suggested for international communication as a means of producing more accessible texts by limiting the vocabulary and the variety of syntactic structures used. Fifty years later, the Plain English Campaign \citep{Crystal-PlainEnglish-1987} offered a Crystal Mark scheme which entails manually checking documents for simplified English. The campaign was followed by releases of several guidelines for producing easy-to-read English texts for people with intellectual disabilities \citep{Freyhoff-98,Making-myself-clear}, 
making public information more accessible by using plain English \citep{Plain-11}, and making web content more accessible \citep{W3C,Cooper&al'10}. 
Most guidelines were initially proposed for English, and later adapted to other languages. For example, Rational French \citep{Barthe&al'1999} was inspired by AECMA Simplified English, a controlled language used in the aerospace industry \citep{stajner-2021-automatic}. 

Presently, easy-to-read news articles are offered on specialized websites in many countries, e.g.\ Noticias f\'{a}cil in Spain\footnote{www.noticiasfacil.es}, DR in Denmark\footnote{https://www.dr.dk/ligetil/}, News Web Easy in Japan\footnote{https://www3.nhk.or.jp/news/easy/}. Several websites, e.g.\ Newsela\footnote{www.newsela.com}, and News in Levels\footnote{https://www.newsinlevels.com/} offer news articles at several complexity levels and are designed for English language learners. Those websites, however, offer only a few current articles at any one time as manual adaptation cannot keep up with new information being released on the web.

\subsection{Lexical Simplification}

Lexical simplification is a natural language processing (NLP) task of automatically simplifying words and phrases in a given text or sentence, so that the information is more understandable to the reader. At the same time, lexical simplification needs to preserve the original meaning of the text or sentence. Depending on the level of simplification applied, some non-essential nuances of the original meaning are allowed to be lost. Nevertheless, the core information needs to be preserved. Due to a high potential of being used for social good  and improving social inclusion of many people, lexical simplification has been attracting growing attention from the NLP community \citep{Saggion-17-book,stajner-2021-automatic}. 

The main reason for focusing on automated tools for lexical simplification is that manual transformation of complex vocabulary into basic vocabulary in a given text is expensive, time-consuming, and requires professional editors. Furthermore, it has been shown that different target groups, e.g.\ native and non-native speakers, need different words to be simplified \citep{YimamEtAl-2017-IJCNLP}. In the case of people with cognitive or reading impairments, needs for vocabulary simplification are even more heterogeneous \citep{OrasanEtAl-18}. In the case of language learners, which words need to be simplified depends not only on their language proficiency level, but also on their native language \citep{aprosio-etal-2018-L2-LS}. Lexical simplification should thus ideally be personalized. Well-performing lexical simplification systems would significantly lower the editing costs and enable publishing more up-to-date articles understandable to wider populations, and offer a possibility for developing personalized readers.

Although it has been attracting the attention of the NLP community since the late 1990s, lexical simplification has only recently gained popularity as the underlying technologies have advanced \citep{stajner-2021-automatic}. The majority of proposed systems focus on the English language \citep{PaetzoldSpecia-LSsurvey-17,Alva-SSsurvey-2020}, as it is the case with many tasks in the NLP world. Several works have proposed lexical simplification systems for Spanish \citep{Bott&al'2012,CASSA-NAACL-15,ferresSG_BGNLP2017}, Portuguese \citep{hartmann-etal-2020-simplex}, French \citep{hmida-etal-2018-assisted}, Chinese \citep{ChineseLS-2021}, Japanese \citep{kajiwara-yamamoto-2015-evaluation,hading-etal-2016-japanese} and Swedish \citep{abrahamsson-etal-2014-medical}. 

Lexical simplification consists of four subtasks that can be modelled either separately (in a modular approach) or jointly (in an end-to-end approach): (1) complex word/phrase identification (CWI or CPI); (2) generation of possible substitutes (SG); (3) selection of substitutes that fit the context and preserve the original meaning (SS); and (4) ranking of substitutes (SR).

The goal of the first subtask is to ensure that only those words and phrases that are difficult for the target reader(s) are simplified. Some works opt for not having this subtask and instead treat all content words as potentially difficult words. In such lexical simplification systems, the other subtasks treat the original/target word as one of the candidate substitutions. Having a Complex Word Identification (CWI) module at the beginning of the lexical simplification pipeline has been shown to improve the performance of lexical simplification systems by avoiding unnecessary errors stemming from trying to simplify words that do not need to be simplified \citep{Paetzol&Specia'2015}. In the NLP community, complex word identification is treated as a separate NLP task and has attracted a lot of attention through two shared tasks: SemEval 2016 CWI for English \citep{Paetzold&Specia'2016a}, and the BEA 2018 CWI shared task for English, German and Spanish, as well as multilingual CWI including English, German, Spanish, and French \citep{yimam-etal-2018-report}. The SemEval 2021 shared task on lexical complexity prediction \citep{shardlow-etal-2021-semeval} also provided a new dataset for complex word and multi-word expressions identification for English. 
In the second sub-task (SG), candidate substitutions are usually retrieved either from specialized dictionaries and thesaurus, or by leveraging statistical properties of large corpora. The approach chosen for this sub-task influences coverage of lexical simplification systems. The third sub-task (SS) is crucial for ensuring that the original meaning has been preserved during lexical simplification, as it checks whether or not the substitutes fit the context well and convey the same semantics as the original word. This sub-task also ensures grammaticality of the output.  
The method used for ranking the simplicity of the substitution candidate, in the fourth sub-task (SR), should be chosen depending on specific lexical simplification needs of the target population or target user. In an end-to-end approach to lexical simplification, this sub-task should ideally be modelled separately from the rest, to allow for adaptation of the system to particular needs of different users or target populations. 

As there are already many high-quality datasets for the evaluation of complex word/phrase identification modules, in this study, we focus on datasets necessary for evaluating the three other aspects of the lexical simplification pipeline (generation of substitution candidates, selection of substitutes that fit the context and preserve the original meaning, and ranking of substitutes according to their simplicity).

\subsection{Evaluation Datasets for Lexical Simplification}

The main bottleneck for building reliable and efficient lexical simplification systems is the absence of datasets for training and evaluation. The absence of datasets for training has been mitigated by using unsupervised methods. Since 2015, lexical simplification systems that use word embedding vectors \citep{Glavas&Stajner'2015,Paetzold-Specia-16-AAAI} and neural language models \citep{shardlow-nawaz-2019-neural,qiang2020BERTLS} dominate the field. The absence of reliable datasets for automatic evaluation is, nevertheless, a great issue, as evaluation with target users requires significant time and specifically trained human assessors, and is thus not optimal for prototyping lexical simplification systems \citep{stajner-2021-automatic}. 

Evaluation datasets for lexical simplification systems only exist for a handful of languages: English \citep{Specia&al'2012,horn2014learning,Paetzold-BenchLS-2016}, French \citep{rolin-etal-2021-frenlys}, Portuguese \citep{Hartmann_Aluisio_2020},  Spanish \citep{Alarcon-21-LSdataset}, Japanese \citep{kajiwara-yamamoto-2015-evaluation}, and Chinese \citep{ChineseLS-2021}. The SemEval-2012 dataset \citep{Specia&al'2012} only evaluates one aspect of English lexical simplification systems, the ranking of the substitution candidates. The other datasets were used to evaluate all aspects of the lexical simplification pipeline: generation of the substitution candidates, their ranking, and fitting to the given context. Of the above-mentioned evaluation datasets, only the English ones have the status of the benchmark datasets. Evaluation datasets for other languages contain a small number of annotations by only one-to-five people. They were thus used only as a proxy for evaluating proposed lexical simplification systems in corresponding languages (see Section~\ref{sec:RW} for more details).

Another issue with the existing evaluation datasets is that they are not comparable across languages due to different procedures used to select and annotate the instances. Therefore, they cannot be used for the evaluation of multilingual lexical simplification systems and better understanding of strengths and weaknesses of different approaches to lexical simplification should they be adapted to other languages. To fill this gap, in this study, we present a multilingual benchmark dataset for evaluation of lexical simplification systems that consists of instances in English, Spanish, and Brazilian Portuguese, all selected and annotated in a comparable manner using best practices.

\subsection{Contributions}

The work presented in this study makes four contributions to the field of lexical simplification:

\begin{enumerate}
    \item Compilation of three comparable benchmark datasets for evaluation of lexical simplification systems for English, Spanish, and Portuguese;
   \item Comparison of several evaluation methods for assessing efficacy of automatic lexical simplification systems and discussion of their strengths and weaknesses; 
    \item Comparison of the performances of state-of-the-art lexical simplification systems for English, Spanish, and Portuguese on the new benchmark datasets;
    \item Detailed description of the process of compiling the benchmark datasets to offer a possibility for compiling comparable datasets in other languages.
\end{enumerate}

\section{Related Work}\label{sec:RW}

In this section, we give an overview of existing lexical simplification systems for English, Spanish, and Portuguese (Section~\ref{sec:SOTA}), evaluation datasets for lexical simplification and their limitations (Section~\ref{sec:EvaluationDatasets}), and commonly used evaluation metrics for measuring performance of lexical simplification systems (Section~\ref{sec:EvaluationRW}). These sections lay out the reasons for the choices made in this work, i.e.\ the choice of lexical simplification systems used in experiments, design of the annotation experiments, and the choice of evaluation metrics used to showcase the usefulness of the newly compiled dataset.

\subsection{State-of-the-Art Lexical Simplification Systems for English, Spanish, and Portuguese}\label{sec:SOTA}

Since 2015 and up to 2020, neural LS systems which leverage word embeddings for the retrieval of the substitution candidates and their ranking \citep{Glavas&Stajner'2015, Paetzold-Specia-16-AAAI} were considered the state-of-the-art for LS in English. Due to their unsupervised nature and the use of word embeddings trained on vast amounts of data, they have significantly better coverage and adaptability than the previously proposed non-neural systems \citep{Biran&al'2011,horn2014learning}. The systems proposed by \citet{Glavas&Stajner'2015} and \citet{Paetzold-Specia-16-AAAI} show similar performances on several benchmarks \citep{PaetzoldSpecia-LSsurvey-17}, the former being computationally lighter as it uses 200-dimensional pretrained word embeddings, while the latter requires training of 1300-dimensional word embeddings. Two other neural systems \citep{paetzold-specia-2017-lexical,gooding-kochmar-2019-recursive} also rank highly on the common LS benchmarks \citep{qiang2020lsbert}, but they are supervised. This lowers their potential to be adapted to other languages which do not offer much training data. 

Currently, the best performing LS system for English is the LSBert system \citep{qiang2020BERTLS}, which uses pre-trained transformer language model BERT \citep{devlin-etal-2019-bert} and a masking technique for finding suitable simplifications for complex words. This approach was further extended by \citet{przybyla-shardlow-2020-multi} to build a multi-word LS system for English. The LSBert system \citep{qiang2020BERTLS} and our adaptation of it to Spanish and Portuguese will be described in more details in Section~\ref{sec:LSBert}.

Lexical simplification in languages other than English attracted less attention. 
For Spanish, several LS systems have been proposed so far:

\begin{itemize}

\item {\bf LexSiS} \citep{Bott&al'2012} -- an unsupervised lexical simplification system for Spanish that uses an online dictionary and Web as a corpus to compute three features (word vector model, word frequency, and word length) for finding the best substitution candidates. Morphological generation of the right inflection for the best substitute is done by a combination of hand-crafted rules and dictionary look-up.

\item {\bf CASSA} \citep{CASSA-NAACL-15} -- an unsupervised lexical simplification approach for Spanish that uses Google Books Ngram Corpus, the Spanish OpenThesaurus, and web frequencies for finding the best substitution candidates. This approach only finds the best lemma and does not perform morphological generation of the right inflection.

\item {\bf TUNER} \citep{ferresAS17} -- an unsupervised lexical simplification approach for  Spanish, Portuguese, Catalan, and Galician. This system achieves the state-of-the-art results in lexical simplification for Spanish. It will be described in more details in Section~\ref{sec:TUNER}, as one of the systems we adapt to all three languages and use in our experiments.

\item {\bf EASIER} \citep{alarconCTTS2021} -- neural lexical simplification systems for Spanish, which leverage pretrained word embedding vectors and BERT models. The systems were evaluated only for three sub-tasks: CWI, SG, and SS. The CWI sub-task was evaluated using the CWI 2018 shared task dataset for Spanish \citep{yimam-etal-2018-report}. The other two sub-tasks (SG and SS) were evaluated using the EASIER-500 corpus \citep{Alarcon-21-Easier500}. The fourth sub-task, ranking of substitutes (SR), was not evaluated as no Spanish lexical simplification datasets existed that could be used for that purpose \citep{alarconCTTS2021}. 
\end{itemize}

For Portuguese (regardless of the language variant), only three systems that perform text simplification were proposed so far. All three were built and evaluated for Brazilian Portuguese. The system proposed by \citet{Specia-10} is a machine translation-based sentence simplification system. It is a fully supervised system that relies on parallel original-simple sentences for training. It performs several transformations at the same time: lexical simplification, word and clause reordering, syntactic simplification, etc. The other two systems \citep{hartmann-etal-2020-simplex,Hartmann_Aluisio_2020} are lexical simplification systems which are particularly designed to simplify texts for children.

To showcase the usability of the Portuguese portion of our dataset, we adapt TUNER and LSBert to (Brazilian) Portuguese.

\subsection{Existing Evaluation Datasets for Lexical Simplification}\label{sec:EvaluationDatasets}

\begin{table}[]
\begin{center}
\scalebox{0.67}{
\begin{tabular}{llrrrrrrrp{5.5cm}}
\toprule
\multirow{2}{*}{Lang.\ } 
& \multirow{2}{*}{Dataset} & \multirow{2}{*}{\#instances} & \multirow{2}{*}{\#targ.} & \multirow{2}{*}{\#syn.} & \multicolumn{4}{c}{\#annotators} & \multirow{2}{*}{annotator type}\\
& & & & & CWI & SG & SS & SR\ & \\
\\ \midrule
EN 
& LSeval \citep{Belder-LexicalTest-2012} & 430 & 43 & 5.04 & 0 & \multicolumn{2}{r}{5 (SG+SS)} & 5 & AMT, US-based, $>$95\% acc.\ rate\\  
EN 
& LexMTurk* \citep{horn2014learning} & 500 & 459 &  12.58 & 0 & \multicolumn{2}{r}{50 (SG+SS)} & 0 & AMT, US-based, $>$95\% acc.\ rate\\

EN 
& CEFR-LS \citep{uchida-etal-2018-cefr} & 406 & 406 &  2.35 & 0 & 0 & 1 & 0 & native English speaker, expert\\
\\ \midrule
PT(BR) 
& SIMPLEX-PB 3.0 \citep{Hartmann_Aluisio_2020} &  1719 & 757 & 7.31 & 0 & \multicolumn{2}{r}{2 (SG+SS)}  & NA & expert linguists\\
\\ \midrule
FR 
& FrenLyS \citep{rolin-etal-2021-frenlys} & 196 & 196 & 4.03 & ? & 0 & 3 & 20 & SG: automatic, SS: expert linguists, SR: native speakers\\
\\ \midrule
ZH 
& HanLS \citep{ChineseLS-2021} & 524 & 524 & 8.51 & 2 &  \multicolumn{2}{r}{5 (SG+SS)} & mix & native speakers\\
\\ \midrule
JP 
&SNOW E4 \citep{kajiwara-yamamoto-2015-evaluation} & 2330 & & 4.50 & 0&1&5&5 & crowdsourced\\
JP & BCCWJ \citep{kodaira-etal-2016-controlled} & 2010 & 210 & 4.30 & 0 & 5 & mix & mix & native, $>$95\% acc.rate\\ 
\\ \midrule
ES & EASIER-500 \citep{Alarcon-21-Easier500} & 500 & 500 & 3 & 1 & \multicolumn{2}{r}{1 (SG+SS)} & NA & expert linguist\\
\hline
\end{tabular}}
\caption{Datasets used in evaluation of lexical simplification systems. Commonly used benchmark datasets are denoted with an `*' next to their name. The column `\#targ.' denotes the total number of unique target words (in many datasets, the same target word is given in 10 different contexts). The column `\#syn.' denotes the average number of unique simpler synonyms proposed by all annotators per instance. The column `\#annotators' denotes the number of annotators that provided their annotations for the specific task: CWI (complex word identification), SG (generation of possible substitutes), SS (substitute selection), and SR (substitute ranking). }\label{tab:LSdatasets}
\end{center}
\end{table}

The main characteristics of the existing datasets for evaluation of lexical simplification systems are given in Table~\ref{tab:LSdatasets}. The following characteristics were taken into account:
\begin{itemize}
    \item \emph{\#instances}: the total number of instances/contexts;
    \item \emph{\#targ.}: total number of target/complex words;
    \item \emph{\#syn.}: the average number of simpler synonyms per target word;
    \item \emph{CWI}: the number of annotators (per instance) who pointed out complex words that need to be simplified;
    \item \emph{SG}:\footnote{In some cases, the same annotators were asked to suggest substitutes that fit in the context (preserve the original meaning and grammaticality). In those cases, the number of annotators is given jointly for columns SG and SS and denoted as (SG+SS).} the number of annotators (per instance) who suggested potential substitutes;
    \item \emph{SS}: the number of annotators (per instance) who selected substitutes (from the list of potential substitutes) that correctly preserve the original meaning and fit in the context (preserve both semantics and grammaticality);
    \item \emph{SR}: the number of annotators (per instance) who ranked the selected substitutes based on their simplicity;
    \item \emph{annotator type}: description of the annotators as given in the respective papers that describe the datasets.
\end{itemize}

The existing datasets all follow different procedures for data selection and annotation. For example, the context sentences have been selected from various genres and topics:
\begin{itemize}
    \item Wikipedia: LexMTurk (EN)
    \item Introduction parts from introductory textbooks on various topics (economics, psychology, sociology, etc.): CEFR-LS (EN)
    \item Internet texts (balanced): LSeval (EN), BCCWJ (JP)
    \item Mixture of textbooks and dialogues for children: SIMPLEX-PB 3.0 (PT-BR), FrenLyS (FR)
    \item Mixture of original and translated texts: HanLS (ZH)
    \item Newspapers: SNOW E4 (JP)
\end{itemize}

The procedures used to choose target words (i.e.\ complex words to be simplified) also vary across the datasets. Most of them use a fully automatic method. Nevertheless, they still differ in which fully automatic method they use. LSeval (EN), CEFR-LS (EN), SIMPLEX-PB 3.0 (PT-BR), BCCWJ (JP) select the target words by leveraging dictionaries of easy/complex words, LexMTurk (EN) by leveraging automatic alignment of original and Simple English Wikipedia, while SNOW E4 (JP) bases the choice of target words on dictionaries of easy words and word frequency counts in newspapers.

For FrenLyS \citep{rolin-etal-2021-frenlys}, instances were selected from two corpora, ALECTOR \citep{gala-etal-2020-alector} and texts from various textbooks. For the instances originating from ALECTOR corpus, complex words were identified based on the information gained from a reading experiment with dyslexic children. For the instances originating from various textbooks, complex words were identified based on a reading experiment with various readers and the recorded reading time. Given that dyslexic readers have different lexical simplification needs than neurotypical readers \citep{Rello&al'2013a,Rello'2014a}, it is important to identify the target audience when working with such corpora.
%it is not clear why were those instances merged into one dataset, and what type of lexical simplification (for whom?) should ideally be evaluated with FrenLyS. 

For obtaining a list of replacement candidates, some datasets relied on human annotation and some on automatic generation of the candidates. For LSeval (EN), LexMTurk (EN), HanLS (ZH), SNOW E4 (JP), candidates were generated and selected jointly, by asking annotators to suggest simpler synonyms that fit the context (by preserving original meaning and grammaticality). For creation of BCCWJ (JP) dataset, one set of annotators suggested replacement candidates, and the other set of annotators assessed whether or not those candidates are a good fit (preserving meaning and grammaticality). For CEFR-LS (EN) dataset, replacement candidates were obtained automatically based on thesaurus and dictionaries. For FrenLyS (FR), replacement candidates were also obtained automatically, but in this case, based on several sources: thesaurus, word embeddings, and neural language models.

The instructions given to annotators for how to select the right candidates, i.e.\ to judge candidate fitness in context also varied across the datasets. During the creation of FrenLyS (FR), the annotators were instructed to judge the candidate substitution correct if it does not change original meaning, and to accept hypernyms and hyponyms, and small changes in nuances as correct \citep{rolin-etal-2021-frenlys}. During the creation of CEFR-LS (EN), in contrast, the annotators were instructed to select the given substitution candidate, only if it successfully conveys the nuance of the target word in the specific context and does not affect the meaning of a sentence \citep{uchida-etal-2018-cefr}. For SNOW E4 (JP) datasets, the annotators were instructed to select the substitution candidate if it sounds natural in the context and does not change the original meaning \citep{kajiwara-yamamoto-2015-evaluation}. In contrast to all other datasets, substitute selection in BCCWJ (JP) was done using majority vote on top of human suggestions. 

The procedures for candidate ranking also varied across the evaluation datasets. LSeval (EN), FrenLys (FR), and SNOW E4 (JP) used human ranking with different type and number of annotators. Candidate ranking in LexMTurk (EN) was automatic, based on the frequency of the candidate being proposed by 50 crowdworkers. In CEFR-LS (EN) dataset, candidates were also ranked automatically, but in this case, based on leveraging special sources (word lists and CEFR language learning framework). In HanLS (ZH) and BCCWJ (JP) datasets, candidates were first ranked by humans, and then the final ranks were computed by using the mean value of the ranks given by each annotator during suggestion of adequate replacement candidates (HanLS), or maximum likelihood estimation on top of five human rankings (BCCWJ).

For all above-mentioned differences in procedures used to select and annotate instances in existing evaluation datasets, it is not possible to compare performances of lexical simplification systems in different languages. Furthermore, some of the datasets offer too few simpler synonyms per target word to make evaluation metrics applied to them reliable. Our dataset, in contrast, uses comparable procedures for selection and annotation of instances across three languages, which makes the results of lexical simplification systems comparable across languages. Due to a high number of annotators per instance (25), our dataset also offers a higher number of simpler synonyms per target word, which results in higher reliability of standard evaluation metrics applied on it.

\subsection{Evaluation Metrics for Lexical Simplification}\label{sec:EvaluationRW}

The common metrics for the evaluation of lexical simplification systems are defined by \citet{Paetzold-BenchLS-2016} as: 
\begin{itemize}
    \item {\bf Potential} -- the percentage of instances for which at least one of the substitutions generated is present in the gold standard;
    \item {\bf Precision} -- the percentage of generated candidates that are in the gold standard;
    \item {\bf Recall} -- the percentage of gold-standard substitutions that are included in the generated substitutions;
    \item {\bf F1} -- the harmonic mean of Precision and Recall.
\end{itemize}

All metrics can be calculated taking into account all outputs of the system for each instance, or only taking the first $k$ (ranked) outputs of the system for each instance in which case they are usually denoted as Potential@K, Precision@k, Recall@k, and F1@k. Using the metrics ``@k" instead of the original ones is recommended when comparing systems of different architectures, especially those that notably differ in the number of simpler substitutes generated per instance. Systems that output a higher number of simpler substitutes would have notably higher Potential and Recall, and lower Precision than the systems that output only a few simpler substitutes.

Each metric aims to evaluate different aspects of lexical simplification systems. Potential is used to evaluate the substitution candidate generation (SG) phase of the systems, i.e.\ the potential of a system to generate at least one simpler substitute. Precision evaluates system's performance at generating and selecting substitution candidates (SG and SS phases). Recall evaluates versatility of generated simpler synonyms. When used only on the $k$ best ranked simpler substitutes, all measures additionally evaluate the ranking capabilities of the system (SR phase).  

Recall, and thus also F1, is additionally influenced by the number of simpler substitutes per instance, and their quality/correctness, in the gold data. Therefore, Recall and F1 are meaningful metrics only in the case of carefully curated gold data, i.e.\ gold data that is known to be easier for the target population. On the benchmark datasets where gold data consists of candidate replacements suggested by crowdsourced workers as simpler synonyms for the given target word in context, Recall and F1 may not be the right evaluation metrics. Such datasets, instead, are better suited for using Potential and Precision, especially if each instance contains more than just a few simpler synonyms suggested (as it is the case in our dataset, see Table~\ref{tab:TSAR-Data} in Section~\ref{sec:DataStats}).

\section{New Evaluation Dataset for Lexical Simplification}

\subsection{\textcolor{black}{Data Collection}}

\textcolor{black}{We compiled a new dataset of examples of lexical simplifications across Portuguese, Spanish and English. Crowdsourced workers were presented with instances (sentences) in which a single token is marked as requiring simplification. They were asked to provide simpler synonyms for the marked words, taking into account that the original meaning of the sentence should be preserved.} 

The following example is taken directly from the English portion of our dataset:

\begin{quote}
    The daily death toll in Syria has declined as the number of \textbf{observers} has risen, but few experts expect the U.N. plan to succeed in its entirety.
\end{quote}

We deliberately present target words in context as the returned words should be grammatically aligned with the context and semantically consistent with the original term. Grammatical alignment is important as the words should be directly replaceable in the original context. If a word is in the wrong tense, or requires a missing preposition, then the resulting sentence would be ungrammatical. Similarly, if a word is a good grammatical fit in the sentence, but is not semantically consistent with the original term, then the resulting sentence will be difficult for readers to understand.

In our selection procedure we only identify single words (as opposed to complex multi-word expressions) for annotation. This simplifies the problem space and makes model input, etc.\ easier to process. Whilst we acknowledge that the problem of simplifying complex multi-word expressions is important, we leave this to a more thorough handling in dedicated work on the subject. Although we did not select for multi-word expressions, we did allow annotators to return multiple words if they could not think of a relevant single-word simplification. This allowed for the insertion of function words, or for compound terms to be returned in the rare cases where this was necessary.

In each of our datasets we have explicitly chosen to identify simplifications for words which are known to be difficult for a reader to understand. Another option would have been to simplify every token in a context, or select tokens at random or according to some heuristic (such as low lexical frequency). If we had selected random tokens, many would have not required simplification --- leading to a waste in annotator effort. If we had used frequency or length, we may have missed words which do not follow these patterns --- leading to a biased dataset. By leveraging human annotations of complexity we are able to explicitly produce simplifications for terms which require it.

\textcolor{black}{All instances used in our dataset (for all three languages) were taken from existing corpora that had information about which words need to be simplified. For English and Spanish, we used the respective portions of the 2018 edition of the Complex Word Identification shared task \citep{yimam-etal-2018-report}. For (Brazilian) Portuguese, we used the PorSimplesSent dataset \citep{leal-etal-2018-nontrivial}. Sentences in all three corpora (English CWI-2018, Spanish CWI-2018, and PorSimplesSent) often contained several words marked for requiring simplification. For the examples in our new dataset, however, we opted for marking only one of those words in each instance. The reason for this was two-fold. First, if we had marked all originally marked complex words in each sentence, the task would be much more complex for crowdsourced workers. Apart from having to propose a simpler synonym that fits well in the context, they would need to pay attention to how all proposed simpler synonyms in a given sentence interact. This would lead to longer cognitive effort by the annotators and higher number of incorrect substitution candidates. It would also make the validation of collected annotations difficult. For example, it could happen that each suggested simpler synonym in a sentence is correct on its own, but together with other suggested simpler synonyms in the sentence it does not sound natural. Second, the state-of-the-art lexical simplification systems only perform simplification of a single complex word in a given sentence at time. If the given sentence contains several words marked as complex, the state-of-the-art lexical simplification systems would simplify them in several rounds, i.e.\ first simplifying one of them, then the next one (this time the context would be different as the first complex word is already replaced), and so on.}

Specific collection protocols for the source data and complex words in each language are described in the subsections below.

\subsubsection{Portuguese}

We extracted instances of Portuguese complex words in context from the PorSimplesSent dataset \citep{leal-etal-2018-nontrivial}. Each instance in PorSimplesSent was collected from Brazilian newspapers and was therefore of the Brazilian Portuguese (PT-BR) variety. The PorSimplesSent dataset consists of a collection of original and simplified sentences, whereby a trained linguist manually simplified a complex sentence \textcolor{black}{according to detailed guidelines}. To extract complex words from this dataset, we conducted automatic word alignment. A script compared each word within the original and simplified sentence pair and identified potential inconsistencies between the two. A native PT-BR linguist then manually examined these inconsistencies and recognized those instances which contained an accurate simplification for a particular target word. These target words were subsequently considered complex. In total, 386 sentence pairs were found to contain a complex word. 348 of these instances contained unique complex words, whereas 39 instances contained a duplicate complex word but in a unique context.

The 386 instances, containing the original complex word in context, were then shared with 25 crowd-sourced MTurk annotators located in Brazil. \textcolor{black}{They were asked to provide the most suitable simplification for each given complex word. This resulted in 9604 suggested simplifications}. Analysis of the provided simplifications found that 70 unique candidate substitutions (10.25\% of all suggested simplifications) were either (a) equal to the complex word, (b) not PT-BR, or (c) inappropriate (e.g. words that did not accurately preserve the meaning of the sentence or the original complex word). These candidate substitutions were excluded resulting in a final total of 2742 unique candidate substitutions, or 8620 repeated simplifications, for 386 instances.

\subsubsection{Spanish}

For Spanish, a set of 588 examples were extracted from the CWI Shared Task 2018 dataset\footnote{{\url{https://sites.google.com/view/cwisharedtask2018/datasets}}}  \citep{yimam-etal-2018-report}. \textcolor{black}{Only the examples with terms annotated as complex by five or more native language annotators were taken into account.} This set was then reduced to 402 examples after a manual judgment process that involved two computational linguistics experts. 

The manual judgment process was conducted to decide if the complex word was ``simplifiable"\footnote{That is if the experts were able to find a simpler substitute for the complex word.} in its context or not. The experts could choose from three options: ``simplifiable", ``not simplifiable", or ``dubious". This resulted in three sets of judgements: 1) a set of 256 examples for which both experts agreed that the complex word is simplifiable, 2) a set of 113 examples for which both experts agreed that the complex word is not simplifiable, and 3) a set of 219 examples for which there is disagreement between the experts, or at least one of the experts indicated that they had doubts about the simplification. This manual judgment process was done using the aid of online dictionaries and thesaurus. \textcolor{black}{Finally, after a joint revision of the 219 previously selected dubious examples, a subset of 146 examples was re-classified as simplifiable, thus leading to a total of 402 simplifiable instances}. 

After deleting repeated examples, a set of 393 unique simplifiable examples was obtained. An additional filtering step was applied afterwards. It involved removing cases of: (1) very similar examples; (2) complex words from other languages that are not yet commonly used and not (yet) accepted as valid words in Spanish (e.g.\ \emph{hoax}); (3) complex words that have a sense in the sentence that is used in very specific locations (e.g. \emph{jirón} - when refers to a sense related with a kind of street). This resulted in a final set of 384 instances for crowdsourced annotation of candidate substitutes for lexical simplification.

\textcolor{black}{For every example a simpler substitute was proposed by a set of 25 annotators (3 splits of 128 instances were submitted to 75 different annotators, 25 annotators per split). The demographics data of the 75 annotators is as follows: {\em Gender}: Female (47), Male (28); {\em Age Ranges (years)}: 20-30 (54)  31-40 (17)  41-50 (2)  50-59 (1), Unknown (1);  {\em Nationality}:  Argentina (1), Greece (1), Italy (2), Venezuela (2), Portugal (4), Spain (6), Chile (13), Mexico (45), Unknown (1). }

Once the crowdsourced annotation process was finished, it was decided to exclude three instances: two instances with the complex word repeated two times in its context, and a sentence which has a typographical error. This resulted in 381 instances in the final dataset. 

The final dataset contains 356 different target words: 333 words appear once, 21 words appear twice, and two words appear three times.
There are a total of 9524 substitutions in the dataset and after joining the repeated substitutions in each instance we get a total of 3918  different substitutions.

One of the authors reviewed the crowdsourced annotations and detected that they contain 137 incorrect substitutions (1.44\%) and 93 dubious substitutions\footnote{Dubious substitutions are those which, according to the reviewer, are probably incorrect.} (0.98\%), 230 substitutions equal to the complex word (2.41\%), and 9064 correct substitutions (95.17\% of the total substitutions). Although these incorrect and dubious substitutions (according to the reviewer) were not excluded from the dataset, we think that this information can be used in further work (in collaboration with linguistic experts) to generate a cleaner set of examples.\footnote{The Spanish part of the dataset has been described in more details in \citep{ferrs-saggion:2022:LREC}.}

\subsubsection{English}

The English data from the 2018 edition of the Complex Word Identification shared task (comprising data from  news, Wikinews and Wikipedia articles) \citep{yimam-etal-2018-report} was selected as the initial set of instances for annotation. This comprised 34879 instances, each of which had binary complexity annotations from 10 native and 10 non-native English speakers. We selected all instances where terms had been annotated as difficult by at least five native annotators. \textcolor{black}{We then removed any duplicate tokens and contexts giving 1949 instances to select from.}

\textcolor{black}{We manually selected 400 instances from the set of 1949 possible candidates. These were selected by a native English speaker and were identified as those instances where a lexical simplification could reasonably be produced, i.e. if the annotator could find at least one single word replacement to simplify the sentence, the instance was kept. If this was not possible, the instance was discarded. Each instance comprised of a token and the context in which that token occurred.}

\textcolor{black}{These instances were passed on to annotation using Amazon's Mechanical Turk. We did not record demographic statistics on the annotators, however we requested annotators from English speaking countries to maintain quality. Each instance was annotated by 25 annotators, each of whom were instructed to return a single word to simplify the sentence.  The native English annotator reviewed all suggestions ($n=10000$) to determine if they were acceptable in the context of the task and removed unsuitable annotations where the guidelines had not been followed (i.e. not simplifying, returning dictionary definitions, etc.). When suggestions were removed further annotations were requested too ensure each instance had 25 suggestions. 14 instances were removed during this process where it was clear that no good suggestions could be found by the crowd workers, leading to 386 final instances.}

\subsection{Data Annotation}

We annotated the data using crowdsourcing platforms (Amazon Mechanical Turk for English and Portuguese, Prolific for Spanish). The Spanish instances were annotated first, and the guidelines presented in that annotation round were translated into English and Portuguese, with minimal editing to ensure that the task remained the same across languages. One notable difference was that whereas Spanish and Portuguese require gender agreement for replacements, this usually does not apply in English. The text used in the guidelines across the three languages is shown in Appendix I. When using crowdsourcing, it is important to keep the guidelines brief to encourage the annotators to read them, whilst also allowing them to gain enough knowledge to complete the task without being rejected.

 We only rejected instances in cases where the data that was returned was clearly abusing the guidelines (e.g., dictionary definitions, whole sentences, nonsense input). All instances were manually verified for correctness and instances were manipulated to fit the context grammatically where necessary. Only affix changes were applied, keeping the original semantics of the simplifications.
 
\subsection{Data Description and Statistics}\label{sec:DataStats}

The data comprises of 1153 instances, split across the three languages. Summary statistics for the data are shown in Table~\ref{tab:TSAR-Data}, and one instance from each language in Table~\ref{tab:sample_dataset}. The data is intended as a benchmark test set for Lexical Simplification systems in one of the languages, and for multilingual systems. The gold annotations consists of all simpler substitutes suggested by crowdsourced workers, checked for quality by at least one computational linguist who is native speaker of the respective language. The suggested simpler synonyms are ordered (in descending order) by the number of annotators who suggested them. 

\begin{table}[]
    \begin{center}
    \scalebox{1.0}{

    \begin{tabular}{l c c c c c r}
    \toprule
        \multirow{2}{*}{Language} & \multirow{2}{*}{Instances} & \multirow{2}{*}{Tokens} & \multirow{2}{*}{Contexts} & \multicolumn{3}{c}{Suggestions}  \\ 
        & & & & Min & Max & Avg \\\\ \midrule
        English (EN) & 386 & 330 & 369 & 2 & 22 & 10.55 \\
        Spanish (ES) & 381 & 356 & 326 & 2 & 19 & 10.28 \\
        (Brazilian) Portuguese (PT-BR) & 386 & 348 & 356 & 1 & 16 & 8.10 \\

        \\ \midrule
        All &1153 & 1031 & 1051 & 1 & 22 & 9.64 \\\hline
        
    \end{tabular}
    }
    \caption{Statistics on the TSAR-ST 2022 Lexical Simplification Dataset. \textcolor{black}{The column `Instances' signifies the total number of instances used in the crowdsourcing lexical simplification experiments, while columns `Tokens' and `Contexts' signify the total number of unique tokens and contexts in those instances, respectively. The columns `Min', `Max', and `Avg' represent the minimal, maximal and average number of unique simpler substitutes suggested by the crowdsourced workers per instance.}}
    \label{tab:TSAR-Data}
\end{center}    
\end{table}

\begin{table}
\begin{center}
\scalebox{0.95}{
\begin{tabular}{lp{8cm}p{7cm}}

\toprule
Lang.\ & Sentence (target word in bold) & Simpler substitutes suggested by 25 crowdsourced workers\\
\\ \midrule

EN & {\it A local witness said a separate group of attackers} {\bf disguised} {\it in burgas -- the head-to-toe robes worn by conservative Afghan women -- then tried to storm the compound}	
& concealed:4, dressed:4, hidden:3, camouflaged:2, changed:2, covered:2, disguised:2, masked:2, unrecognizable:2, converted:1, impersonated:1\\
 \\ \midrule
ES & {\it Conforme avanzaba el debate en el Congreso de Filadelfia, Lee iba asumiento una posici\'{o}n m\'{a}s favorable a la independencia total y no s\'{o}lo a la autonom\'{i}a del Imperio Brit\'{a}nico, su} {\bf convicci\'{o}n} {\it de la necesidad de la independencia logr\'{o} convencer a delegados de otras colonias e incluso persuadi\'{o} a sus propios electores de Virginia, temerosos que Lee pudiera llegar demasiado lejos.} & creencia:5, seguridad:5, certeza:5, convencimiento:3, ideal:2, f\'{e}:1, persuaci\'{o}n:1, fuerte creencia:1, idea:1\\
\\ \midrule
PT-BR & {\it Quem n\~{a}o conseguir} {\bf esgotar} {\it o armazenamento de diesel puro n\~{a}o pode misturar com o b2 porque o produto ficaria fora de especifica\c{c}\~{a}o.} & acabar:10, esvaziar:7, acabar com:4, gastar:1, consumir:1, diminuir:1, zerar:1\\
\\ \midrule
\end{tabular}
}
 \caption{Examples of instances from the dataset (the number after `:' in the third column represents the number of crowdsourced workers that suggested that replacement).}
\label{tab:sample_dataset}
\end{center}
\end{table}

Table \ref{tab:TSAR-Data} shows that the dataset is evenly divided between the three languages that we have chosen to annotate. In each language, we started off with 400 instances, but discarded instances that were troublesome for annotators, leaving 386 instances in English and Portuguese and 381 instances in Spanish. The number of unique tokens and contexts is close to the total number of instances, indicating that there is little repetition of tokens across the dataset. Whilst repeated instances may be interesting to explore the effect of context on polysemy and replacement fit, the repeated tokens will likely have similar replacements, in a similar order. For this reason they were kept to a minimum. No two instances are identical (e.g.\ having the same token and context). The number  of suggestions varies similarly across data subsets, with one or two suggestions being offered in the minimal case and up to 22 suggestions being offered for the largest case. On average, there are 9.64 suggestions returned per instance in our dataset. The most frequent suggestion typically far outnumbered the long tail of other, less popular suggestions.

Lexical simplification systems typically employ a natural language engineering approach, leveraging state of the art technology, rather than simply training systems to directly perform simplification. Because of this,  our data is intended as a benchmarking test set and not for training. We expect that systems will leverage other resources to improve their performance. Therefore, we have not split our data into training and test subsets.

A number of domains are represented in our dataset, resulting from the diverse corpora that were used to select contexts for each language. The English data was selected from the CWI-2018 dataset, containing Wikitext and news data. The Spanish data was selected also from the CWI-2018 dataset for Spanish, which contains data from the Spanish Wikipedia. The Portuguese data was taken from general and scientific news articles.

We paid annotators at the following rates per instance in each language: \$0.03 for English, \$0.14 for Spanish and \$0.02 for Portuguese. An additional \$0.01 was paid in platform fees per instance for English and Portuguese. In  total, this equates to a spend of \$2481.73 to annotate our entire dataset.

We can easily add in further languages at a future point by running further annotations with the same protocol. For languages where a CWI dataset exists, we can use the protocol employed for English and Spanish, whereas for languages without CWI data, we can follow the same protocol as for Portuguese.

\subsection{Limitations of the Dataset}

Although of high quality and being the only multilingual evaluation dataset for lexical simplification, our dataset has some notable limitations.

One limitation of the English and Spanish portion of the dataset is that all instances come from the same source, covering only one genre. Therefore, they only provide a reliable evaluation of lexical simplification systems which focus on those specific genres.

Another limitation of our dataset is that the provided replacements represent simpler synonyms according to the crowdsourced workers, rather than experts in the area. The high number of annotators per instance (25) mitigates this issue to some extent, as it offers a possibility for ranking the replacement candidates according to the number of times they were suggested by different people. The most frequently proposed replacement candidates could thus be considered of a good quality, but some re-ranking may be required to confirm this.

\subsection{Intended Use}\label{sec:IntendedUse}

The new dataset is envisioned as the first evaluation dataset that allows for fair comparison of lexical simplification systems across different languages (English, Spanish, and Portuguese), due to comparable procedures for selecting and annotating instances in all three languages. As it contains a high number of simpler replacements suggested for each target/complex word, it is particularly valuable for evaluating substitution generation (SG) and substitution selection (SS) modules of lexical simplification systems. Due to a high number of annotations per instance (25 crowdsourced workers), it can also be used to evaluate substitute ranking (SR) capabilities of lexical simplification systems. However, as gold data was crowdsourced, and only professionally checked for preservation of grammaticality and original meaning, the ranking of substitution candidates can only be used as a proxy for general notion of simplicity, and not as simplicity ranking for any particular target user/population.

Given a high number of simpler replacements offered for each target/complex word, this dataset can also be used as a starting point for building evaluation datasets for lexical simplification intended for some particular audience. In that case, the original set of simpler replacements should be filtered and ranked based on expert annotations (e.g.\ by carers or expert psycholinguists aware of particular simplification needs of the target user(s)) or user studies with the target users (e.g.\ comprehension tests, eye-tracking studies, etc.).

\section{Experiments}

Our newly compiled evaluation dataset for lexical simplification allows us to compare the performance of different lexical simplification approaches across the three languages we have incorporated (English, Spanish, and Portuguese). To demonstrate this, we adapt the state-of-the-art lexical simplification system for Spanish (TUNER) and the state-of-the-art lexical simplification system for English (LSBert) to all three languages (English, Spanish, and Portuguese). As well as representing state-of-the-art systems in two languages, these systems also represent two different approaches to lexical simplification. TUNER relies on static resources such as vocabularies and thesauri, whereas LSBert relies on a large scale BERT-based language model. 

We further compare the performances of those systems on our benchmark dataset using several evaluation metrics that aim to capture different aspects of system's performances. The next three subsections describe: (1) the TUNER lexical simplification system (a non-neural system) and its adaptation to the three languages (Section~\ref{sec:TUNER}); (2) the LSBert lexical simplification system (a system that leverages neural language models) and its adaptation to the three languages (Section~\ref{sec:LSBert}); and  (3) the evaluation metrics used for comparing performances of TUNER and LSBert across the three languages (Section~\ref{sec:EvaluationMetricsUsed}). 

\subsection{TUNER-LS}\label{sec:TUNER}

The TUNER Candidate Ranking System used in this evaluation is an adaptation of the TUNER Lexical Simplification architecture \citep{ferresSG_BGNLP2017} to work with Spanish, Portuguese and English. Some components for English were obtained from the YATS Simplifier for English \citep{ferresMSA16}. The TUNER simplifies words (common nouns, verbs, adjectives, and  adverbs) in context. The adaptation presented here omits the Complex Word Identification phase and the Context Adaptation phase, returning the lists of ranked candidates with correct inflections instead of returning the complete sentence simplified using the top ranked candidate.

The adapted system has the following phases (executed sequentially): (1) Sentence Analysis, (2) Word Sense Disambiguation (WSD), (3) Synonyms Ranking, and (4) Morphological Generation. The Sentence Analysis phase uses the FreeLing 4.0 system  to perform tokenization, sentence
splitting, part-of-speech (PoS) tagging,  lemmatization, and Named Entity Recognition.

The WSD algorithm used is based on the Vector Space Model approach for lexical semantics. The thesauri used for WSD were extracted from FreeLing 4.0 data which is derived from Multilingual Central Repository (MCR) 3.0\footnote{\url{http://adimen.si.ehu.es/web/MCR/}} (release 2012). 
Each thesaurus contains a set of synonyms and its associated set of senses with related synonyms (see the number of entries and senses of the thesaurus for each language used in Table \ref{wsdresources}). The WSD algorithm uses a word vector model derived from a large text collection from which a vector for each word in the thesaurus is created by collecting co-occurring word lemmas of the word in  11-word window (five content words to each side of the target word) contexts (only nouns, verbs, adjectives, and adverbs). Then, a common vector is computed for each of the word senses of a given target word (lemma and PoS) by adding the vectors of all words in each sense.  When a complex word is detected, the WSD algorithm computes the cosine distance between the context vector computed from the words of the complex word context (at sentence level) and the
word vectors of each sense from the model. The word sense selected is
the one with the lowest cosine distance between its word vector in the
model and the context vector of the complex word in the sentence or
document to simplify.

\begin{table}
\begin{center}

\begin{tabular}{lccrrccrr}
\toprule
            
\multirow{2}{*}{Language} && &\multicolumn{2}{c}{EuroWordNet} & && \multicolumn{2}{c}{Wikipedia\tablefootnote{We used the Simple English Wikipedia for English (EN)}}  \\
&&& \#entries & \#senses & &&  \#documents & \#words  \\
\\ \midrule
               EN & && 63649 & 87792 & && 99943 & 15M\\
 ES  & &&  36571  & 50397 & && 1061535 & 349M \\
 PT  & && 35635 & 45737  & && 956553  & 203M \\
 
\\ \midrule
\end{tabular}
\caption{Statistics of the EuroWordNet thesaurus and the Wikipedia collections processed for the TUNER-LS system.}
	\label{wsdresources}
\end{center}
\end{table}

The word vector models for each language were extracted from Wikipedia dumps. For Spanish and Portuguese, the Spanish Wikipedia and Portuguese Wikipedia were used, respectively. For English, the word vectors model was extracted from the Simple English Wikipedia.\footnote{Simple Spanish Wikipedia and Simple Portuguese Wikipedia do not exist.} The plain text of the  documents was extracted (see in Table \ref{wsdresources} the number of documents and words extracted from the Wikipedia dump). The FreeLing 3.1 system was used to extract the lemmas and PoS tags of each word, from an 11-word window (five content words to each side of the target word).

The Synonym Ranking phase ranks synonyms using word form (or lemma) frequency as a simplicity measure. The frequency list used by each of the languages are: 1) Spanish Wikipedia form counts for Spanish, 2) Portuguese Wikipedia form counts for Portuguese, and 3) Simple English Wikipedia word form counts for English.

The Morphological Generator module generates the correct inflected forms
of the final selected synonyms. Given a set of pairs $<\text{LEMMA}, \text{PoS}>$, with the lemma corresponding to a substitution candidate and the PoS tag corresponding to the PoS Tag of the complex word in the sentence, this module returns the inflected forms of the candidates. For Spanish and Portuguese this module uses an algorithm that combines lexicon-based generation and predictions from decision-trees (see \citep{ferresAS17} for a more detailed description of this system).   For English, the SimpleNLG Java API \footnote{https://github.com/simplenlg/simplenlg} was used with its default lexicon to perform this task using the candidate and and the PoS tag of the complex word in the original sentence.

\subsection{LSBert}\label{sec:LSBert}

The LSBert\footnote{\url{https://github.com/qiang2100/BERT-LS}} \citep{qiang2020lsbert} state-of-the-art lexical simplifier for English has been adapted to deal also with Spanish and Portuguese. The LSBert system uses a pretrained representation of BERT to propose substitution candidates with high grammatical and semantic similarity to a complex word in a sentence. LSBert uses the masked language model (MLM) of BERT to predict a set of candidate substitution words and their substitution probabilities. BERT is fed with the original sentence concatenated with a copy of the sentence in which the complex word has
been masked. LSBert combines five different features for Lexical Simplicity Ranking: BERT prediction order, a BERT-based language model, the PPDB database, word frequency, and
word semantic similarity with fasttext.

In the original LSBert system the simplification algorithm selects the top ranked candidate and performs the simplification only if the top candidate has a higher frequency (frequency feature) or lower loss (language model feature) with respect to the original complex word, otherwise returns the same complex word. In our adaptation, we modified the system to retrieve up to $K=5$ candidates from a BERT-based pre-trained model in this way: after the ranking procedure the top $K=5$ candidates that a have higher frequency (frequency feature) or have a lower loss (language model feature) with respect to the complex word are selected.

For English, we used the same resources described in the original LSBert paper \citep{qiang2020lsbert}: (1) BERT-large WWM, (2) Porter Stemmer, (3) Fasttext CBOW model for English\footnote{\url{https://dl.fbaipublicfiles.com/fasttext/vectors-english/crawl-300d-2M-subword.zip}}, (4) SUBTLEX zipf values \cite{Brysbaert2009MovingBK} and (5) PPDB 2.0\footnote{\url{http://nlpgrid.seas.upenn.edu/PPDB/eng/ppdb-2.0-tldr.gz}}. The language specific resources used  to adapt the system to Spanish and Portuguese were: (1) BERT-base based models: BETO \citep{CaneteCFP2020} for Spanish and BERTimbau for Portuguese \citep{souza2020bertimbau}, (2) Snowball stemmer for Spanish and Portuguese, (3) Fasttext CBOW model for  Spanish\footnote{\url{https://dl.fbaipublicfiles.com/fasttext/vectors-crawl/cc.es.300.vec.gz}} and Portuguese\footnote{\url{https://dl.fbaipublicfiles.com/fasttext/vectors-crawl/cc.pt.300.vec.gz}}, and (4) SUBTLEX-ESP (Spanish) and SUBTLEX-PT (Portuguese) word frequencies in zipf format. 
The configurations parameters used were: probability-mask=0.5, max-sequence-length=350.
    
\begin{table}[]
    \begin{center}
    \scalebox{1.0}{

    \begin{tabular}{llllcc}
    \toprule
            
Language & Model & Type & Case & Layers & \#parameters  \\
\\ \midrule
               English & BERT-WWM & BERT-large-wwm & uncased &24 & 336M \\
Spanish  &  BETO  & BERT-base-wwm & uncased &12& 110M \\
(Brazilian) Portuguese  &  BERTimbau & BERT-base & cased &  12 & 110M \\
 
\\ \midrule
		\end{tabular}}
	\caption{Features of the BERT-based models used in the LSBert approach for each language.}
		\label{featuresBERTmodels}
\end{center}
\end{table}

\subsection{Evaluation Metrics}\label{sec:EvaluationMetricsUsed}

We evaluate TUNER and LSBert lexical simplification systems for English, Spanish, and Portuguese on the respective portions of our dataset using \textbf{Potential}, \textbf{Precision} and \textbf{Recall}.
As mentioned in Section~\ref{sec:EvaluationRW}, those metrics are the usual metrics for evaluation of lexical simplification systems. For a more detailed comparison of the systems across languages and architecture, we use Potential@1,\footnote{Note that Precision@1 is equal to Potential@1.} Potential@3, Potential@5, Precision@3, Precision@5, and Recall@5. In all cases, Precision and Recall are computed for each instance and then averaged over all instances in the respective portion of the dataset.

We additionally define and calculate {\bf Accuracy@1}  -- the percentage of instances for which the best ranked substitution generated by the system is the same as the most frequently suggested simpler synonym in the gold data. The rationale for introducing this metric is to be able to evaluate the possibility of using state-of-the-art lexical simplification systems in a fully automatic lexical simplification setup, where presented with a sentence and a target word in it, the system automatically generates a new sentence with a correct replacement. To evaluate this scenario, we assume that the most frequently proposed simpler replacement in the gold data corresponds to a surely correct simplification in the given context.

\section{Results and Discussion}

Table~\ref{tab:Results} presents results of LSBert and TUNER models (in all three languages) on the new dataset. Several instances together with their gold annotations and outputs of LSBert and TUNER systems are presented in Tables~\ref{tab:OutputEN}--~\ref{tab:OutputPT} in Appendix II.

\begin{table}[]
    \begin{center}
    \scalebox{1.0}{

    \begin{tabular}{llrrrrrrrrr}
    \toprule
       \multirow{2}{*}{System} & \multirow{2}{*}{Language} & \multicolumn{3}{c}{Potential} & \multicolumn{3}{c}{Precision} & \multicolumn{2}{c}{Recall} & \multirow{2}{*}{Accuracy@1}\\
       & & @1 & @3 & @5 & @1 & @3 & @5 & @3 & @5 & \\
\\ \midrule

\multirow{3}{*}{LSBert} & EN & 60.1 & 82.4 & 87.8 & 60.1 & 47.0 & 37.0 & 16.8 & 21.6 & 30.8\\
& ES & 28.9 & 49.3 & 61.4 & 28.9 & 23.3 & 19.6 & 8.4 & 11.7 & 9.7\\
& PT-BR & 32.4 & 50.0 & 58.3 & 32.4 & 23.2 & 18.0 & 9.6 & 12.3 & 15.5\\
\\ \midrule
\multirow{3}{*}{TUNER} & EN & 24.6 & 42.0 & 44.0 & 24.6 & 19.4 & 18.3 & 6.1 &  7.2 & 10.9\\
& ES & 8.9 & 13.9 & 14.4 & 8.9 & 6.6 & 6.3 & 2.5 & 2.7 & 5.5\\
& PT-BR & 17.3 & 26.9 & 26.9 & 17.3 & 12.7 & 12.0 & 5.1 & 5.3 & 10.6\\
\hline      
    \end{tabular}
    }
    \caption{Performances of TUNER and LSBert lexical simplification systems on the new dataset (note that Potential@1 and Precision@1 give the same results as per their definitions).}
    \label{tab:Results}
\end{center}    
\end{table}

\subsection{Comparison of Metrics}

We have presented four metrics for the evaluation of lexical simplification and evaluated two systems across three languages using these metrics, as shown in Table~\ref{tab:Results}. These metrics allow us to better understand the nature of the lexical simplification task and how to evaluate it well.

 \begin{figure}[h!]
 \centering
\includegraphics[width=0.65\textwidth, angle=0]{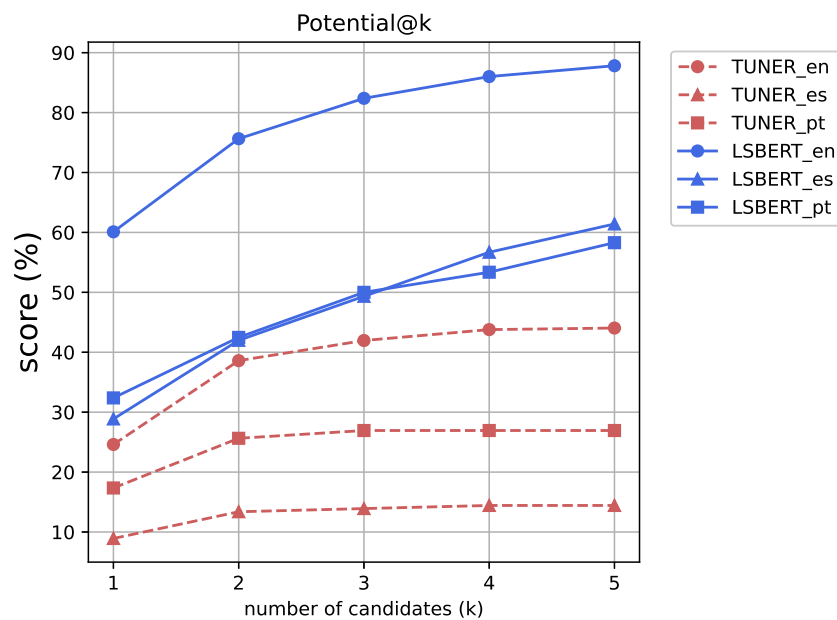}
\caption{This graphic shows the Potential@k from k=\{1,2,3,4,5\} for all the systems and languages evaluated.}
\label{fig1} 
 \end{figure}
 
 \begin{figure}[h!]
\centering
\parbox{8cm}{
\includegraphics[width=9cm]{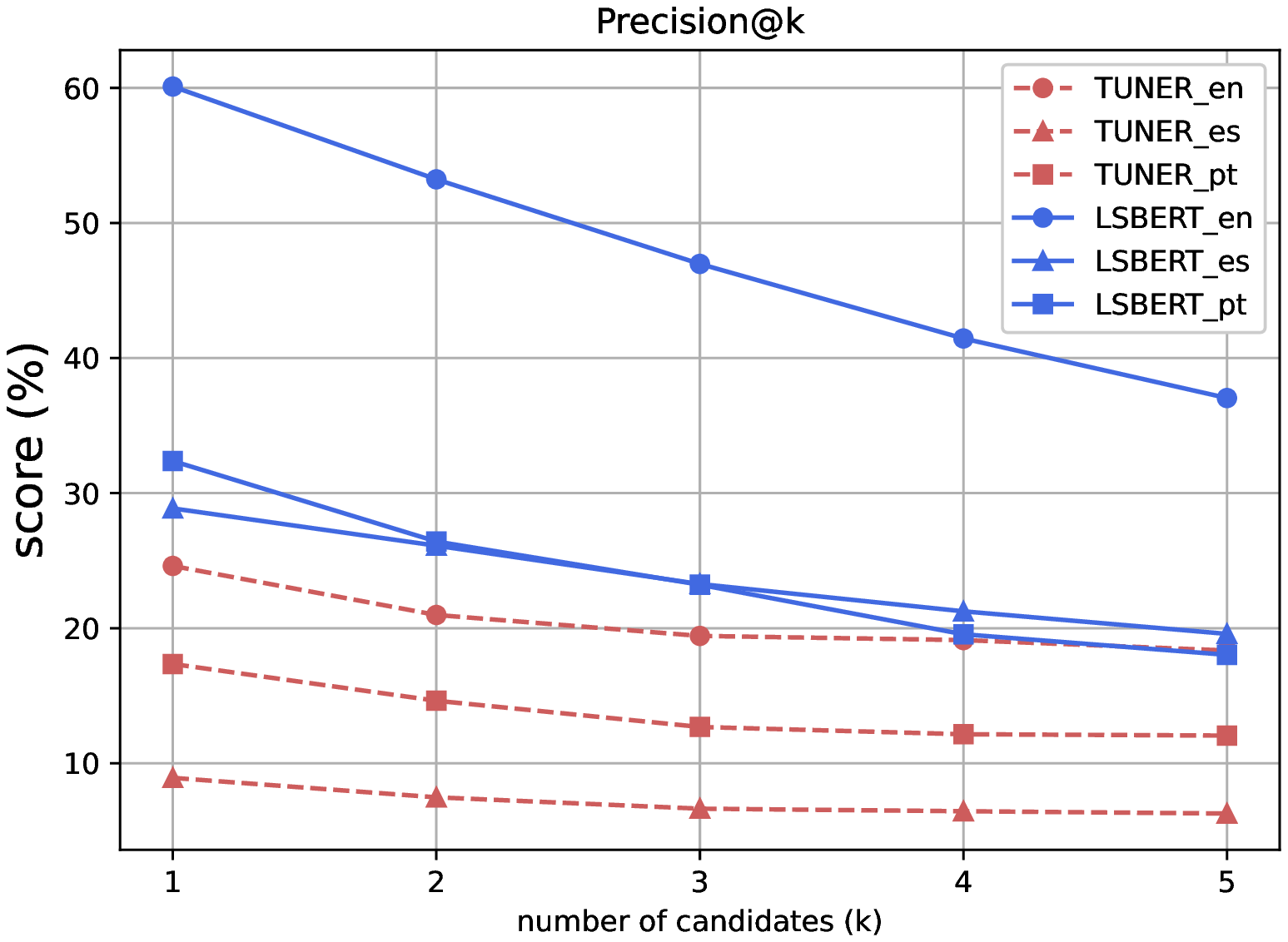}
\caption{Precision@k, k=\{1,2,3,4,5\}.}
\label{fig:2figsA}}
\qquad
\begin{minipage}{8cm}
\includegraphics[width=9cm]{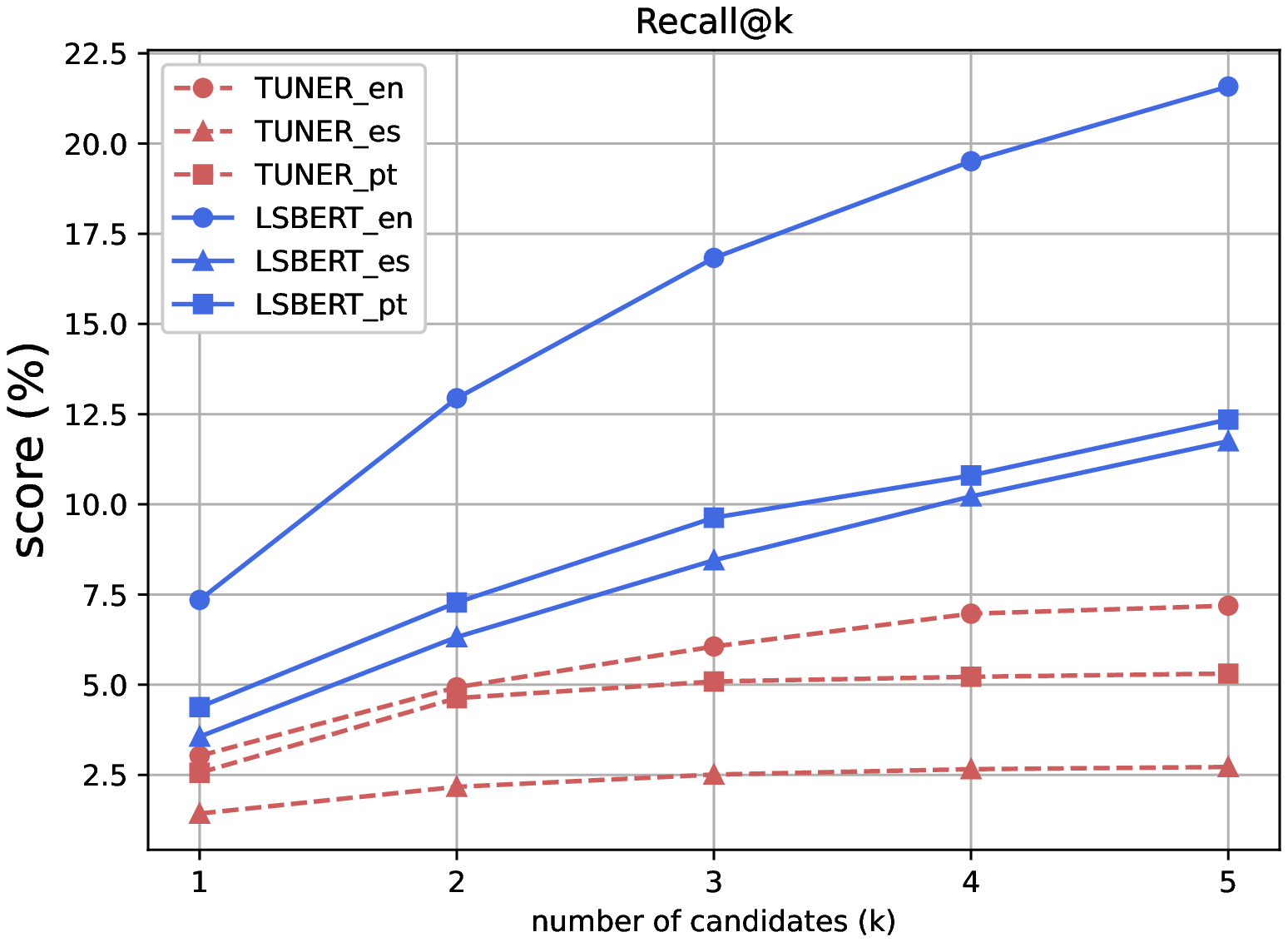}
\caption{Recall@k, k=\{1,2,3,4,5\}}
\label{fig:2figsB}
\end{minipage}
\end{figure}

Our first metric, Potential (visualised in Figure \ref{fig1}), is a very relaxed metric that indicates whether any suggested candidate can be found in the gold standard. Systems can do artificially well on this metric by proposing many unrelated candidates (or indeed an entire vocabulary) and so we limit our evaluation to the first 1, 3 and 5 candidates proposed by a system. In this setting, we are able to attain scores that indicate that systems are usually able to suggest at least one relevant suggestion, with the probability being higher if more candidates are allowed.

Unlike Potential, Precision decreases as the number of candidates that are considered increases (Figure~\ref{fig:2figsA}). A Precision@1 of 60.1 for LSBert in English indicates that the system is usually suggesting a word that is among the gold standard annotations. Precision drops as K increases because each candidate considered must be available in the gold standard. The candidates are returned ranked by likelihood of inclusion, so considering further less likely candidates is bound to lead to fewer inclusions for each instance.

Recall (Figure \ref{fig:2figsB}) is consistently the lowest metric across our evaluation. We previously discussed (Section~\ref{sec:EvaluationRW}) the suitability of Recall  when the gold data contains a large number of simpler substitutes per instance, as in our dataset. We thus report the values for Recall@3 and Recall@5 here only to have complete benchmark results. Recall is consistently the lowest metric, and is heavily dependent on the number of instances in the gold data for each instance.

We also considered a new metric for simplification which is Accuracy@1. This replicates the simplification scenario of an automated system that is selecting a simplification for replacement in a sentence. This metric requires the system to choose exactly the same candidate as the top-ranked candidate in the gold standard. Our best system (LS-BERT on English) achieved this 30.8\% of the time. This only indicates to us that the system replicated the top choice of the gold standard at the given rate. It may have been the case that in the 69.2\% of remaining cases another valid, but less likely candidate was chosen. This has some crossover with our potential metric, that shows that as the number of candidates considered increases the likelihood of finding a valid candidate also increases. Our Accuracy@1 metric is helpful to show how well our systems can perform in a very strict setting, replicating the annotators decisions, but should always be taken in the context of other metrics that give a different view of the systems' performance. 

For comparing lexical simplification performances of the systems with different architectures and across the three languages, when using our dataset, we suggest using Potential@1, Potential@3, and Accuracy@1. Potential@1 and Accuracy@1 can be seen as the upper and lower measure of the usefulness of the system which automatically replaces the target word with a simpler (best ranked) synonym. Potential@3 is a valuable metric if the envisioned real-world usage of the system is to suggest a few simpler synonyms to the human editor, and thus lower the editing costs by speeding up manual simplification.

\begin{table}[]
    \begin{center}
    \scalebox{1.0}{

    \begin{tabular}{llrrr}
    \toprule
       \multirow{2}{*}{Source} & \multirow{2}{*}{Language} & \multicolumn{3}{c}{Substitutes}
       \\
       & & Min & Max & Avg\\
\\ \midrule

\multirow{3}{*}{Gold} & English & 2 & 22 & 10.55\\
& Spanish & 2 & 19 & 10.28\\
& Portuguese & 1 & 16 & 8.10\\
\\ \midrule
\multirow{3}{*}{LSBert} & English & 5 & 5 & 5.00\\
& Spanish & 5 & 5 & 5.00\\
& Portuguese & 5 & 5 & 5.00\\
\\ \midrule
\multirow{3}{*}{TUNER} & English & 1 & 12 & 2.99\\
& Spanish & 1 & 12 & 1.59\\
& Portuguese & 1 & 10 & 1.80\\
\hline      
    \end{tabular}
    }
    \caption{The statistics for the number of simpler substitutes in gold data and the systems' output.}
    \label{tab:NumberSubstitutes}
\end{center}    
\end{table}

\subsection{Cross-System Comparisons}

The neural lexical simplification system (LSBert) outperforms the non-neural system (TUNER) in all three languages by all evaluation metrics (Table \ref{tab:Results}). This is not surprising given that LSBert is the state-of-the-art system for lexical simplification in English and it uses resources with better coverage than TUNER. This results in a higher number of final simplification suggestions generated by LSBert. While LSBert generates more simplification suggestions, in this work, we take into account only its first five (best ranked) simplification suggestions as we keep in mind real-world usage and want to have a fairer comparison between LSBert and TUNER. We allow TUNER to vary the number of substitutes returned according to the probabilities of the candidates
words. Nevertheless, although for some instances TUNER was able to generate up to 12 candidates, for the majority of instances TUNER generated fewer candidates than this, with an average number of candidates between 1.59 and 2.99 (Table \ref{tab:NumberSubstitutes}). The gold standard data typically contained more candidates than those proposed by either of the systems. 

Appendix II contains Tables~\ref{tab:OutputEN}~--~\ref{tab:OutputPT}, which demonstrate sample outputs of both systems in each language. LSBert seems to suggest more generalised contextual fits, whereas TUNER appears to suggest more conservative semantically accurate candidates. For example, the following sentence is presented to both systems:

\begin{quote}
    War \textbf{maniacs}
    of the South Korean puppet military made another grave provocation to the DPRK in the central western sector of the front on Thursday afternoon.
\end{quote}

In this example, \textbf{maniacs} is the complex word requiring simplification. LSBert suggests: \textit{criminals, victims, machines, freaks, people,} whereas TUNER suggests:   \textit{lunatics, madmans, maniacs}. The LSBert outputs fit well with the context, but are sometimes semantically incorrect (e.g.\ `war machines'). Another candidate `victims' is in fact the antonym of the original candidate (`maniacs'). TUNER on the other hand proposes three suggestions of reasonable quality. `madmans' is obviously the result of a pipeline error in the morphological adapter. `madman' has been incorrectly pluralised and the correct form should be `madmen'. The relationship of these candidates to the gold standard data is presented in Table \ref{tab:OutputEN}.

A further manual inspection of the systems' output revealed that TUNER is capable to outperform LSBert in some special cases, where the target word appears in a less commonly used context. In such cases, we find that LSBert suggests more frequently used words, but at the cost of severely changing the original meaning of the sentence.

\subsection{Cross-Lingual Comparisons}

We can make two key observations comparing the LS approaches across the three languages.
Firstly, both architectures perform significantly better for English than for the other two languages (Table~\ref{tab:Results}).
Secondly,  the TUNER architecture performs better for Portuguese than for Spanish, by all evaluation metrics (Table~\ref{tab:Results}) and the LSBert architecture performs similar in Portuguese and Spanish with the exception of the Accuracy@1 metric.

These facts may be explained by the following factors: (1) Linguistic differences between English and Romance languages and between Spanish and Portuguese, (2) Dataset specific differences, and (3) The tools and resources used in the simplification algorithms (which all have better performances and coverage for English than for the other two languages). It is reasonable to assume the hypothesis that linguistic differences among Spanish, Portuguese and English can influence the results of our experiments.  Spanish and Portuguese are part of the Ibero-Romance sub-family of Romance languages and English is part of the Germanic family of languages. According to Ethnologue\footnote{\url{www.ethnologe.com}} lexical similarity\footnote{Ethnologue's definition of Lexical Similarity: the percentage of lexical similarity between two linguistic varieties is determined by comparing a set of standardized wordlists and counting those forms that show similarity in both form and meaning.} between Spanish and Portuguese is about 89\%.

On the other hand, although the procedures to collect the subsets for each language were very similar, minor differences in both the level of lexical complexity of complex words and the gold annotated substitutions in each language-specific subset of the trilingual dataset could have influenced the results. The instructions were translated into each language and adapted for language specific concerns (e.g., inflection handling). It is possible that the translated guidelines led to different interpretations of the task, which could affect the internal consistency of each dataset. Further, the annotator pools in each language were selected according to those crowd workers available at annotation time. These groups are mutually exclusive as they were selected for their first language, and so it may be the case that one group returned more or less consistent annotations than the others.

Moreover, the two different algorithms (and associated resources) have 
a notable influence in the differences of results among the three languages tested. For the LSBert algorithm we have several resources to compare among the three languages: BERT model, word-embeddings, stemmer, PPDB, and frequency files.  The BERT model is the most important feature as it is used in the Substitution Generation phase and the Substitution Ranking phase. Regarding the comparison among language-specific BERT models in the experiments reported in Section 4, we have used BERT-large-uncased-WWM for English, BERTimbau BERT-base cased for Portuguese, and BETO BERT-base uncased for Spanish. The size of BERT (BERT-large vs BERT-base) seems not to be the most important factor for those differences, as  in follow-up experiments with LSBERT with BERT-base-uncased for English we obtained a Potential@1 of 0.544 and Accuracy@1 of 0.251, thus still outperforming the results for Portuguese and Spanish. Those results indicate that the BERT model in English has some attributes that greatly outperform the other models in Spanish and Portuguese. 

For the TUNER architecture we can compare the following resources: NLP resources, thesaurus, context-vectors, and frequency counts lists. The most important and influential resources are the thesaurus and the context-vectors. The number of thesaurus entries and senses is notably greater in English with respect to Spanish and Portuguese (Table ~\ref{wsdresources}). This obviously can have a big impact in the results and this might be additionally reflected in the number of generated simpler substitutes (Table~\ref{tab:NumberSubstitutes}). Moreover, the context-vectors are crucial to select the correct set of synonyms. For English we used the Simple Wikipedia, which has less data but uses Simple English words and grammar, and for Spanish and Portuguese we used the original Wikipedia (as Simple Spanish Wikipedia and Simple Portuguese Wikipedia do not exist).

\section{Conclusion}

In this article, we presented a new evaluation dataset for lexical simplification and benchmarked the state-of-the-art lexical simplification systems for English, Spanish, and Portuguese. This dataset has several advantages over the existing evaluation datasets for lexical simplification:
\begin{itemize}
\item It is the first multilingual evaluation dataset, with instances in Portuguese, Spanish, and English, selected and annotated using comparable procedures. As such, it is the first dataset that offers a reliable comparison of system's performances across the three languages.
\item Simpler replacements that preserve the original meaning and grammaticality of the sentence were suggested by 25 people per each target word, resulting in 10 simpler replacements for each target word across all three languages on average.
\item The quality of crowdsourced replacement suggestions was checked by at least one native computational linguist in each language.
\end{itemize}

Due to a large number of simpler replacements per target word (10 on average), the new dataset offers a possibility for further adaptation to evaluation of lexical simplification systems intended for specific target audiences, by ranking the substitutes based on their simplicity for that specific audience.

To demonstrate the usefulness of the new dataset, we adapted the state-of-the-art neural (LSBert) and non-neural (TUNER) lexical simplification systems to all three languages and evaluated them on this dataset. We found that LSBert architecture outperforms TUNER architecture for all three languages (English, Spanish, and Portuguese). We also found that the performance of LSBert significantly drops when the system is adapted to Spanish and Portuguese. Taking into account that neither Spanish or Portuguese are low-resource languages, this finding poses a question about applicability of LSBert method to other languages with even less resources with satisfying performances.

\subsection{Future Work} \label{sec:st}

\subsubsection{Shared Task}

The new multilingual dataset for lexical simplification presented in this study will be used for the shared task organized as a part of the TSAR workshop at the EMNLP 2022 conference. The data will be released as part of the shared task and is split into trial and testing data. The systems that participate in the shared task will be evaluated with metrics used in this work. After the completion of the shared task, the dataset will be further cleaned and enriched based on the manual error analysis performed on the output of the competing lexical simplification systems.

\subsubsection{Dataset Extension}

Our data is constrained to the genres that are represented in the original source corpora. Whilst these corpora are intended for general audiences (comprising of texts that are not specific to any one domain), they still represent the specific instance of style and form that is found in those corpora. Further  work could replicate the process for selecting and annotating instances in specific domains, allowing the creation of LS systems in those domains. For example, medical text could be annotated and selected for medical complex words and then simpler alternatives suggested for these. In domain-specific research, it is important to consult domain-experts for annotation to ensure that the original complex words are properly understood and transformed.

We have incorporated three languages, according to the expertise found in our research team. We would welcome the addition  of further languages to our corpus following the same annotation protocol. For languages where no existing CWI resource already exists, researchers can follow the selection protocol used for Portuguese.

\subsection{Improved Performance}

Our results are intended to represent a strong baseline of performance. We expect that through the shared task we will see further systems that improve on these results, advancing the state of the art in lexical simplification.  Our dataset is released as a test dataset only, and so we expect the majority of systems to be unsupervised in nature. Further work to produce additional resources for training lexical simplification systems will clearly help to  further push the state of  the art, although we expect that this task will remain a hybrid, rather than fully unsupervised task due to the complex nature of the pipeline operations required.

A further use of the dataset is in training multilingual models for lexical simplification. A multilingual model configured to  work  well for the languages in our dataset may also be able to perform  simplification in a zero-shot setting for unseen languages that can be incorporated into the multilingual model.

\section*{Funding}
%Details of all funding sources should be provided, including grant numbers if applicable. Please ensure to add all necessary funding information, as after publication this is no longer possible.
Horacio Saggion and Daniel Ferr\'es   acknowledge support from the project Context-aware  Multilingual Text Simplification (ConMuTeS)  PID2019-109066GB-I00/AEI/10.13039/501100011033 awarded by  Ministerio de Ciencia, Innovación y Universidades (MCIU) and  by Agencia Estatal de Investigación (AEI) of Spain. 

\section*{Acknowledgments}
We would like to thank Kim Cheng Sheang (Universitat Pompeu Fabra) for his assistance with the review of the evaluation software.
%This is a short text to acknowledge the contributions of specific colleagues, institutions, or agencies that aided the efforts of the authors.

\section*{Data Availability Statement}

The datasets can be found at: https://github.com/LaSTUS-TALN-UPF/TSAR-2022-Shared-Task

%The datasets [GENERATED/ANALYZED] for this study can be found in the [NAME OF REPOSITORY] [LINK].
% Please see the availability of data guidelines for more information, at https://www.frontiersin.org/about/author-guidelines#AvailabilityofData

%TC:ignore

\bibliographystyle{Frontiers-Harvard}
\bibliography{test}

\begin{thebibliography}{62}
\providecommand{\natexlab}[1]{#1}
\expandafter\ifx\csname urlstyle\endcsname\relax
  \providecommand{\doi}[1]{doi:\discretionary{}{}{}#1}\else
  \providecommand{\doi}{doi:\discretionary{}{}{}\begingroup
  \urlstyle{rm}\Url}\fi
\providecommand{\selectlanguage}[1]{\relax}
\providecommand{\bibAnnoteFile}[1]{%
  \IfFileExists{#1}{\begin{quotation}\noindent\textsc{Key:} #1\\
  \textsc{Annotation:}\ \input{#1}\end{quotation}}{}}
\providecommand{\bibAnnote}[2]{%
  \begin{quotation}\noindent\textsc{Key:} #1\\
  \textsc{Annotation:}\ #2\end{quotation}}

\bibitem[{Abrahamsson et~al.(2014)Abrahamsson, Forni, Skeppstedt, and
  Kvist}]{abrahamsson-etal-2014-medical}
Abrahamsson, E., Forni, T., Skeppstedt, M., and Kvist, M. (2014).
\newblock Medical text simplification using synonym replacement: Adapting
  assessment of word difficulty to a compounding language.
\newblock In \emph{Proceedings of the 3rd Workshop on Predicting and Improving
  Text Readability for Target Reader Populations ({PITR})} (Gothenburg, Sweden:
  Association for Computational Linguistics), 57--65.
\newblock \doi{10.3115/v1/W14-1207}
\bibAnnoteFile{abrahamsson-etal-2014-medical}

\bibitem[{Alarcon(2021)}]{Alarcon-21-LSdataset}
[Dataset] Alarcon, R. (2021).
\newblock Dataset of sentences annotated with complex words and their synonyms
  to support lexical simplification.
\newblock \doi{10.17632/ywhmbnzvmx.2}
\bibAnnoteFile{Alarcon-21-LSdataset}

\bibitem[{Alarc{\'{o}}n et~al.(2021)Alarc{\'{o}}n, Moreno, and
  Mart{\'{\i}}nez}]{alarconCTTS2021}
Alarc{\'{o}}n, R., Moreno, L., and Mart{\'{\i}}nez, P. (2021).
\newblock {Exploration of Spanish Word Embeddings for Lexical Simplification}.
\newblock In \emph{Proceedings of the First Workshop on Current Trends in Text
  Simplification {CTTS} 2021)} (CEUR-WS.org), vol. 2944 of \emph{{CEUR}
  Workshop Proceedings}
\bibAnnoteFile{alarconCTTS2021}

\bibitem[{Alarcon et~al.(2021)Alarcon, Moreno, and
  Martínez}]{Alarcon-21-Easier500}
Alarcon, R., Moreno, L., and Martínez, P. (2021).
\newblock Lexical simplification system to improve web accessibility.
\newblock \emph{IEEE Access} 9, 58755--58767.
\newblock \doi{10.1109/ACCESS.2021.3072697}
\bibAnnoteFile{Alarcon-21-Easier500}

\bibitem[{Alva-Manchego et~al.(2020)Alva-Manchego, Scarton, and
  Specia}]{Alva-SSsurvey-2020}
Alva-Manchego, F., Scarton, C., and Specia, L. (2020).
\newblock Data-driven sentence simplification: Survey and benchmark.
\newblock \emph{Computational Linguistics} 46, 135--187
\bibAnnoteFile{Alva-SSsurvey-2020}

\bibitem[{Aprosio et~al.(2018)Aprosio, Menini, Tonelli, Ducceschi, and
  Herzog}]{aprosio-etal-2018-L2-LS}
Aprosio, A.~P., Menini, S., Tonelli, S., Ducceschi, L., and Herzog, L. (2018).
\newblock {Towards Personalised Simplification based on L2 Learners’ Native
  Language}.
\newblock In \emph{{Proceedings of the Fifth Italian Conference on
  Computational Linguistics (CLIC-IT)}}. 305--310
\bibAnnoteFile{aprosio-etal-2018-L2-LS}

\bibitem[{Baeza-Yates et~al.(2015)Baeza-Yates, Rello, and
  Dembowski}]{CASSA-NAACL-15}
Baeza-Yates, R., Rello, L., and Dembowski, J. (2015).
\newblock {CASSA: A Context-Aware Synonym Simplification Algorithm}.
\newblock In \emph{Proceedings of the 2015 Conference of the North American
  Chapter of the Association for Computational Linguistics: Human Language
  Technologies} (ACL), 1380--1385
\bibAnnoteFile{CASSA-NAACL-15}

\bibitem[{Barthe et~al.(1999)Barthe, Juaneda, Leseigneur, Loquet, Morin,
  Escande et~al.}]{Barthe&al'1999}
Barthe, K., Juaneda, C., Leseigneur, D., Loquet, J.-C., Morin, C., Escande, J.,
  et~al. (1999).
\newblock {{GIFAS} Rationalized French: A Controlled Language for Aerospace
  Documentation in French.}
\newblock \emph{Technical Communication} 46, 220--229
\bibAnnoteFile{Barthe&al'1999}

\bibitem[{Biran et~al.(2011)Biran, Brody, and Elhadad}]{Biran&al'2011}
Biran, O., Brody, S., and Elhadad, N. (2011).
\newblock {Putting it Simply: a Context-Aware Approach to Lexical
  Simplification}.
\newblock In \emph{Proceedings of the 49th Annual Meeting of the Association
  for Computational Linguistics} (Portland, Oregon, {USA}), ACL, 496--501
\bibAnnoteFile{Biran&al'2011}

\bibitem[{Bott et~al.(2012)Bott, Rello, Drndarevi\'{c}, and
  Saggion}]{Bott&al'2012}
Bott, S., Rello, L., Drndarevi\'{c}, B., and Saggion, H. (2012).
\newblock {Can Spanish Be Simpler? LexSiS: Lexical Simplification for Spanish}.
\newblock In \emph{Proceedings of the 24th International Conference on
  Computational Linguistics (COLING)}. 357--374
\bibAnnoteFile{Bott&al'2012}

\bibitem[{Brysbaert and New(2009)}]{Brysbaert2009MovingBK}
Brysbaert, M. and New, B. (2009).
\newblock {Moving beyond Kučera and Francis: A critical evaluation of current
  word frequency norms and the introduction of a new and improved word
  frequency measure for American English}.
\newblock \emph{Behavior Research Methods} 41, 977--990
\bibAnnoteFile{Brysbaert2009MovingBK}

\bibitem[{Ca\~{n}ete et~al.(2020)Ca\~{n}ete, Chaperon, Fuentes, Ho, Kang, and
  P{\'{e}}rez}]{CaneteCFP2020}
Ca\~{n}ete, J., Chaperon, G., Fuentes, R., Ho, J.-H., Kang, H., and
  P{\'{e}}rez, J. (2020).
\newblock {Spanish Pre-Trained BERT Model and Evaluation Data}.
\newblock In \emph{PML4DC at ICLR 2020}
\bibAnnoteFile{CaneteCFP2020}

\bibitem[{Cooper et~al.(2010)Cooper, Reid, Vanderheiden, and
  Caldwell}]{Cooper&al'10}
Cooper, M., Reid, L., Vanderheiden, G., and Caldwell, B. (2010).
\newblock {Understanding WCAG 2.0. A guide to understanding and implementing
  Web Content Accessibility Guidelines 2.0.}
\newblock World Wide Web Consortium (W3C)
\bibAnnoteFile{Cooper&al'10}

\bibitem[{Crystal(1987)}]{Crystal-PlainEnglish-1987}
Crystal, D. (1987).
\newblock \emph{The Cambridge Encyclopedia of Language} (Cambridge University
  Press)
\bibAnnoteFile{Crystal-PlainEnglish-1987}

\bibitem[{De~Belder and Moens(2012)}]{Belder-LexicalTest-2012}
De~Belder, J. and Moens, M.-F. (2012).
\newblock A dataset for the evaluation of lexical simplification.
\newblock In \emph{Computational Linguistics and Intelligent Text Processing},
  ed. A.~Gelbukh (Berlin, Heidelberg: Springer Berlin Heidelberg), 426--437
\bibAnnoteFile{Belder-LexicalTest-2012}

\bibitem[{Devlin et~al.(2019)Devlin, Chang, Lee, and
  Toutanova}]{devlin-etal-2019-bert}
Devlin, J., Chang, M.-W., Lee, K., and Toutanova, K. (2019).
\newblock {BERT}: Pre-training of deep bidirectional transformers for language
  understanding.
\newblock In \emph{Proceedings of the 2019 Conference of the North {A}merican
  Chapter of the Association for Computational Linguistics: Human Language
  Technologies, Volume 1 (Long and Short Papers)} (Minneapolis, Minnesota:
  Association for Computational Linguistics), 4171--4186.
\newblock \doi{10.18653/v1/N19-1423}
\bibAnnoteFile{devlin-etal-2019-bert}

\bibitem[{Ferr{\'{e}}s et~al.(2017)Ferr{\'{e}}s, AbuRa'ed, and
  Saggion}]{ferresAS17}
Ferr{\'{e}}s, D., AbuRa'ed, A., and Saggion, H. (2017).
\newblock {Spanish Morphological Generation with Wide-Coverage Lexicons and
  Decision Trees}.
\newblock \emph{Procesamiento del Lenguaje Natural} 58, 109--116
\bibAnnoteFile{ferresAS17}

\bibitem[{Ferr{\'{e}}s et~al.(2016)Ferr{\'{e}}s, Marimon, Saggion, and
  AbuRa'ed}]{ferresMSA16}
Ferr{\'{e}}s, D., Marimon, M., Saggion, H., and AbuRa'ed, A. (2016).
\newblock {{YATS:} Yet Another Text Simplifier}.
\newblock In \emph{{NLDB}} (Springer), vol. 9612 of \emph{Lecture Notes in
  Computer Science}, 335--342
\bibAnnoteFile{ferresMSA16}

\bibitem[{Ferr{\'e}s et~al.(2017)Ferr{\'e}s, Saggion, and
  G{\'o}mez~Guinovart}]{ferresSG_BGNLP2017}
Ferr{\'e}s, D., Saggion, H., and G{\'o}mez~Guinovart, X. (2017).
\newblock {An Adaptable Lexical Simplification Architecture for Major
  {I}bero-{R}omance Languages}.
\newblock In \emph{Proceedings of the First Workshop on Building Linguistically
  Generalizable {NLP} Systems} (Copenhagen, Denmark: Association for
  Computational Linguistics), 40--47.
\newblock \doi{10.18653/v1/W17-5406}
\bibAnnoteFile{ferresSG_BGNLP2017}

\bibitem[{Ferrés and Saggion(2022)}]{ferrs-saggion:2022:LREC}
Ferrés, D. and Saggion, H. (2022).
\newblock {ALEXSIS: A Dataset for Lexical Simplification in Spanish}.
\newblock In \emph{Proceedings of the Language Resources and Evaluation
  Conference} (Marseille, France: European Language Resources Association),
  3582--3594
\bibAnnoteFile{ferrs-saggion:2022:LREC}

\bibitem[{Freyhoff et~al.(1998)Freyhoff, Hess, Kerr, Tronbacke, and Van
  Der~Veken}]{Freyhoff-98}
Freyhoff, G., Hess, G., Kerr, L., Tronbacke, B., and Van Der~Veken, K. (1998).
\newblock \emph{Make it {S}imple, {E}uropean {G}uidelines for the {P}roduction
  of {E}asy-to{R}ead {I}nformation for {P}eople with {L}earning {D}isability}.
\newblock ILSMH European Association, Brussels
\bibAnnoteFile{Freyhoff-98}

\bibitem[{Gala et~al.(2020)Gala, Tack, Javourey-Drevet, Fran{\c{c}}ois, and
  Ziegler}]{gala-etal-2020-alector}
Gala, N., Tack, A., Javourey-Drevet, L., Fran{\c{c}}ois, T., and Ziegler, J.~C.
  (2020).
\newblock {Alector: A Parallel Corpus of Simplified French Texts with
  Alignments of Misreadings by Poor and Dyslexic Readers}.
\newblock In \emph{Proceedings of the 12th Language Resources and Evaluation
  Conference} (Marseille, France: European Language Resources Association),
  1353--1361
\bibAnnoteFile{gala-etal-2020-alector}

\bibitem[{Glava\v{s} and \v{S}tajner(2015)}]{Glavas&Stajner'2015}
Glava\v{s}, G. and \v{S}tajner, S. (2015).
\newblock {Simplifying Lexical Simplification: Do We Need Simplified Corpora?}
\newblock In \emph{Proceedings of the 53rd Annual Meeting of the Association
  for Computational Linguistics and the 7th International Joint Conference on
  Natural Language Processing of the Asian Federation of Natural Language
  Processing}. ACL, 63--68
\bibAnnoteFile{Glavas&Stajner'2015}

\bibitem[{Gooding and Kochmar(2019)}]{gooding-kochmar-2019-recursive}
Gooding, S. and Kochmar, E. (2019).
\newblock Recursive context-aware lexical simplification.
\newblock In \emph{Proceedings of the 2019 Conference on Empirical Methods in
  Natural Language Processing and the 9th International Joint Conference on
  Natural Language Processing (EMNLP-IJCNLP)} (Hong Kong, China: Association
  for Computational Linguistics), 4853--4863.
\newblock \doi{10.18653/v1/D19-1491}
\bibAnnoteFile{gooding-kochmar-2019-recursive}

\bibitem[{Hading et~al.(2016)Hading, Matsumoto, and
  Sakamoto}]{hading-etal-2016-japanese}
Hading, M., Matsumoto, Y., and Sakamoto, M. (2016).
\newblock {J}apanese lexical simplification for non-native speakers.
\newblock In \emph{Proceedings of the 3rd Workshop on Natural Language
  Processing Techniques for Educational Applications ({NLPTEA})}. 92--96
\bibAnnoteFile{hading-etal-2016-japanese}

\bibitem[{Hartmann et~al.(2020)Hartmann, Paetzold, and
  Alu{\'\i}sio}]{hartmann-etal-2020-simplex}
Hartmann, N., Paetzold, G.~H., and Alu{\'\i}sio, S. (2020).
\newblock {SIMPLEX}-{PB} 2.0: A reliable dataset for lexical simplification in
  {B}razilian {P}ortuguese.
\newblock In \emph{Proceedings of the The Fourth Widening Natural Language
  Processing Workshop}. 18--22.
\newblock \doi{10.18653/v1/2020.winlp-1.6}
\bibAnnoteFile{hartmann-etal-2020-simplex}

\bibitem[{Hartmann and Aluísio(2020)}]{Hartmann_Aluisio_2020}
Hartmann, N.~S. and Aluísio, S.~M. (2020).
\newblock {Adaptação Lexical Automática em Textos Informativos do Português
  Brasileiro para o Ensino Fundamental}.
\newblock \emph{Linguamática} 12, 3--27.
\newblock \doi{10.21814/lm.12.2.323}
\bibAnnoteFile{Hartmann_Aluisio_2020}

\bibitem[{Hmida et~al.(2018)Hmida, Billami, Fran{\c{c}}ois, and
  Gala}]{hmida-etal-2018-assisted}
Hmida, F., Billami, M.~B., Fran{\c{c}}ois, T., and Gala, N. (2018).
\newblock Assisted lexical simplification for {F}rench native children with
  reading difficulties.
\newblock In \emph{Proceedings of the 1st Workshop on Automatic Text Adaptation
  ({ATA})} (Tilburg, the Netherlands: Association for Computational
  Linguistics), 21--28.
\newblock \doi{10.18653/v1/W18-7004}
\bibAnnoteFile{hmida-etal-2018-assisted}

\bibitem[{Horn et~al.(2014)Horn, Manduca, and Kauchak}]{horn2014learning}
Horn, C., Manduca, C., and Kauchak, D. (2014).
\newblock {Learning a Lexical Simplifier Using Wikipedia}.
\newblock In \emph{Proceedings of ACL (Short Papers)}. 458--463
\bibAnnoteFile{horn2014learning}

\bibitem[{Kajiwara and Yamamoto(2015)}]{kajiwara-yamamoto-2015-evaluation}
Kajiwara, T. and Yamamoto, K. (2015).
\newblock Evaluation dataset and system for {J}apanese lexical simplification.
\newblock In \emph{Proceedings of the {ACL}-{IJCNLP} 2015 Student Research
  Workshop} (Beijing, China: Association for Computational Linguistics),
  35--40.
\newblock \doi{10.3115/v1/P15-3006}
\bibAnnoteFile{kajiwara-yamamoto-2015-evaluation}

\bibitem[{Kodaira et~al.(2016)Kodaira, Kajiwara, and
  Komachi}]{kodaira-etal-2016-controlled}
Kodaira, T., Kajiwara, T., and Komachi, M. (2016).
\newblock Controlled and balanced dataset for {J}apanese lexical
  simplification.
\newblock In \emph{Proceedings of the {ACL} 2016 Student Research Workshop}
  (Berlin, Germany: Association for Computational Linguistics), 1--7.
\newblock \doi{10.18653/v1/P16-3001}
\bibAnnoteFile{kodaira-etal-2016-controlled}

\bibitem[{Leal et~al.(2018)Leal, Duran, and
  Alu{\'\i}sio}]{leal-etal-2018-nontrivial}
Leal, S.~E., Duran, M.~S., and Alu{\'\i}sio, S.~M. (2018).
\newblock A nontrivial sentence corpus for the task of sentence readability
  assessment in {P}ortuguese.
\newblock In \emph{Proceedings of COLING}. 401--413
\bibAnnoteFile{leal-etal-2018-nontrivial}

\bibitem[{Mencap(2002)}]{Making-myself-clear}
Mencap (2002).
\newblock \emph{Am I making myself clear? Mencap's guidelines for accessible
  writing}
\bibAnnoteFile{Making-myself-clear}

\bibitem[{OECD(2013)}]{OECD-13}
OECD (2013).
\newblock \emph{{OECD Skills Outlook 2013: First Results from the Survey of
  Adult Skills}}.
\newblock Tech. rep., OECD Publishing
\bibAnnoteFile{OECD-13}

\bibitem[{Ogden(1937)}]{ogden37}
Ogden, C.~K. (1937).
\newblock \emph{{Basic English: A General Introduction with Rules and Grammar}}
  (London: Paul Treber)
\bibAnnoteFile{ogden37}

\bibitem[{Or{\u{a}}san et~al.(2018)Or{\u{a}}san, Evans, and
  Mitkov}]{OrasanEtAl-18}
Or{\u{a}}san, C., Evans, R., and Mitkov, R. (2018).
\newblock \emph{Intelligent Text Processing to Help Readers with Autism} (Cham:
  Springer International Publishing).
\newblock 713--740.
\newblock \doi{10.1007/978-3-319-67056-0_33}
\bibAnnoteFile{OrasanEtAl-18}

\bibitem[{Paetzold and Specia(2015)}]{Paetzol&Specia'2015}
Paetzold, G. and Specia, L. (2015).
\newblock {{LEX}enstein: A Framework for Lexical Simplification}.
\newblock In \emph{Proceedings of the ACL-IJCNLP 2015 System Demonstrations}
  (Beijing, China), 85--90
\bibAnnoteFile{Paetzol&Specia'2015}

\bibitem[{Paetzold and Specia(2016{\natexlab{a}})}]{Paetzold-BenchLS-2016}
Paetzold, G. and Specia, L. (2016{\natexlab{a}}).
\newblock {Benchmarking Lexical Simplification Systems}.
\newblock In \emph{Proceedings of the Tenth International Conference on
  Language Resources and Evaluation (LREC)}. 3074--3080
\bibAnnoteFile{Paetzold-BenchLS-2016}

\bibitem[{Paetzold and Specia(2016{\natexlab{b}})}]{Paetzold&Specia'2016a}
Paetzold, G. and Specia, L. (2016{\natexlab{b}}).
\newblock {{S}em{E}val 2016 {T}ask 11: Complex Word Identification}.
\newblock In \emph{Proceedings of the 10th International Workshop on Semantic
  Evaluation} (San Diego, California: Association for Computational
  Linguistics), SEMEVAL, 560--569
\bibAnnoteFile{Paetzold&Specia'2016a}

\bibitem[{Paetzold and
  Specia(2017{\natexlab{a}})}]{paetzold-specia-2017-lexical}
Paetzold, G. and Specia, L. (2017{\natexlab{a}}).
\newblock Lexical simplification with neural ranking.
\newblock In \emph{Proceedings of the 15th Conference of the {E}uropean Chapter
  of the Association for Computational Linguistics: Volume 2, Short Papers}
  (Valencia, Spain: Association for Computational Linguistics), 34--40
\bibAnnoteFile{paetzold-specia-2017-lexical}

\bibitem[{Paetzold and Specia(2016{\natexlab{c}})}]{Paetzold-Specia-16-AAAI}
Paetzold, G.~H. and Specia, L. (2016{\natexlab{c}}).
\newblock Unsupervised lexical simplification for non-native speakers.
\newblock In \emph{Proceedings of the 30th AAAI}
\bibAnnoteFile{Paetzold-Specia-16-AAAI}

\bibitem[{Paetzold and Specia(2017{\natexlab{b}})}]{PaetzoldSpecia-LSsurvey-17}
Paetzold, G.~H. and Specia, L. (2017{\natexlab{b}}).
\newblock A survey on lexical simplification.
\newblock \emph{Journal of Artificial Intelligence Research} 60, 549–593
\bibAnnoteFile{PaetzoldSpecia-LSsurvey-17}

\bibitem[{PlainLanguage(2011)}]{Plain-11}
[Dataset] PlainLanguage (2011).
\newblock Federal plain language guidelines
\bibAnnoteFile{Plain-11}

\bibitem[{Przyby{\l}a and Shardlow(2020)}]{przybyla-shardlow-2020-multi}
Przyby{\l}a, P. and Shardlow, M. (2020).
\newblock Multi-word lexical simplification.
\newblock In \emph{Proceedings of the 28th International Conference on
  Computational Linguistics} (Barcelona, Spain (Online): International
  Committee on Computational Linguistics), 1435--1446.
\newblock \doi{10.18653/v1/2020.coling-main.123}
\bibAnnoteFile{przybyla-shardlow-2020-multi}

\bibitem[{Qiang et~al.(2020{\natexlab{a}})Qiang, Li, Yi, Yuan, and
  Wu}]{qiang2020BERTLS}
Qiang, J., Li, Y., Yi, Z., Yuan, Y., and Wu, X. (2020{\natexlab{a}}).
\newblock Lexical simplification with pretrained encoders.
\newblock \emph{Thirty-Fourth AAAI Conference on Artificial Intelligence} ,
  8649–--8656
\bibAnnoteFile{qiang2020BERTLS}

\bibitem[{Qiang et~al.(2020{\natexlab{b}})Qiang, Li, Zhu, Yuan, and
  Wu}]{qiang2020lsbert}
Qiang, J., Li, Y., Zhu, Y., Yuan, Y., and Wu, X. (2020{\natexlab{b}}).
\newblock {LSBert: A Simple Framework for Lexical Simplification}.
\newblock \emph{arXiv preprint arXiv:2006.14939}
\bibAnnoteFile{qiang2020lsbert}

\bibitem[{Qiang et~al.(2021)Qiang, Lu, Li, Yuan, and Wu}]{ChineseLS-2021}
Qiang, J., Lu, X., Li, Y., Yuan, Y., and Wu, X. (2021).
\newblock Chinese lexical simplification.
\newblock \emph{IEEE/ACM Transactions on Audio, Speech, and Language
  Processing} 29, 1819--1828.
\newblock \doi{10.1109/TASLP.2021.3078361}
\bibAnnoteFile{ChineseLS-2021}

\bibitem[{Rello(2014)}]{Rello'2014a}
Rello, L. (2014).
\newblock \emph{{DysWebxia. A Text Accessibility Model for People with
  Dyslexia}}.
\newblock Ph.D. thesis, Universitat Pompeu Fabra, Barcelona, Spain
\bibAnnoteFile{Rello'2014a}

\bibitem[{Rello et~al.(2013)Rello, Baeza{-}Yates, Dempere{-}Marco, and
  Saggion}]{Rello&al'2013a}
Rello, L., Baeza{-}Yates, R.~A., Dempere{-}Marco, L., and Saggion, H. (2013).
\newblock {Frequent Words Improve Readability and Short Words Improve
  Understandability for People with Dyslexia}.
\newblock In \emph{Proceedings of the International Conference on
  Human-Computer Interaction (Part IV)} (Cape Town, South Africa), INTERACT,
  203--219
\bibAnnoteFile{Rello&al'2013a}

\bibitem[{Rolin et~al.(2021)Rolin, Langlois, Watrin, and
  Fran{\c{c}}ois}]{rolin-etal-2021-frenlys}
Rolin, E., Langlois, Q., Watrin, P., and Fran{\c{c}}ois, T. (2021).
\newblock {F}ren{L}y{S}: A tool for the automatic simplification of {F}rench
  general language texts.
\newblock In \emph{Proceedings of the International Conference on Recent
  Advances in Natural Language Processing (RANLP 2021)} (Held Online: INCOMA
  Ltd.), 1196--1205
\bibAnnoteFile{rolin-etal-2021-frenlys}

\bibitem[{Saggion(2017)}]{Saggion-17-book}
Saggion, H. (2017).
\newblock \emph{Automatic text simplification} (Morgan \& Claypool Publishers)
\bibAnnoteFile{Saggion-17-book}

\bibitem[{Shardlow et~al.(2021)Shardlow, Evans, Paetzold, and
  Zampieri}]{shardlow-etal-2021-semeval}
Shardlow, M., Evans, R., Paetzold, G.~H., and Zampieri, M. (2021).
\newblock {S}em{E}val-2021 task 1: Lexical complexity prediction.
\newblock In \emph{Proceedings of the 15th International Workshop on Semantic
  Evaluation (SemEval-2021)} (Online: Association for Computational
  Linguistics), 1--16.
\newblock \doi{10.18653/v1/2021.semeval-1.1}
\bibAnnoteFile{shardlow-etal-2021-semeval}

\bibitem[{Shardlow and Nawaz(2019)}]{shardlow-nawaz-2019-neural}
Shardlow, M. and Nawaz, R. (2019).
\newblock Neural text simplification of clinical letters with a domain specific
  phrase table.
\newblock In \emph{Proceedings of the 57th Annual Meeting of the Association
  for Computational Linguistics} (Florence, Italy: Association for
  Computational Linguistics), 380--389.
\newblock \doi{10.18653/v1/P19-1037}
\bibAnnoteFile{shardlow-nawaz-2019-neural}

\bibitem[{Souza et~al.(2020)Souza, Nogueira, and Lotufo}]{souza2020bertimbau}
Souza, F., Nogueira, R., and Lotufo, R. (2020).
\newblock {BERTimbau: Pretrained BERT Models for Brazilian Portuguese}.
\newblock In \emph{Intelligent Systems}, eds. R.~Cerri and R.~C. Prati
  (Springer International Publishing), 403--417
\bibAnnoteFile{souza2020bertimbau}

\bibitem[{Specia(2010)}]{Specia-10}
Specia, L. (2010).
\newblock {Translating from complex to simplified sentences}.
\newblock In \emph{Proceedings of the 9th international conference on
  Computational Processing of the Portuguese Language (PROPOR)} (Springer
  Berlin Heidelberg), vol. 6001 of \emph{Lecture Notes in Computer Science},
  30--39
\bibAnnoteFile{Specia-10}

\bibitem[{Specia et~al.(2012)Specia, Jauhar, and Mihalcea}]{Specia&al'2012}
Specia, L., Jauhar, S.~K., and Mihalcea, R. (2012).
\newblock {SemEval-2012 task 1: English Lexical Simplification}.
\newblock In \emph{Proceedings of the First Joint Conference on Lexical and
  Computational Semantics} (Stroudsburg, PA, USA: Association for Computational
  Linguistics), SemEval, 347--355
\bibAnnoteFile{Specia&al'2012}

\bibitem[{Temnikova et~al.(2015)Temnikova, Castillo, and
  Vieweg}]{temnikova2015emterms}
Temnikova, I.~P., Castillo, C., and Vieweg, S. (2015).
\newblock Emterms 1.0: A terminological resource for crisis tweets.
\newblock In \emph{Proceedings of the ISCRAM}
\bibAnnoteFile{temnikova2015emterms}

\bibitem[{Uchida et~al.(2018)Uchida, Takada, and Arase}]{uchida-etal-2018-cefr}
Uchida, S., Takada, S., and Arase, Y. (2018).
\newblock {CEFR}-based lexical simplification dataset.
\newblock In \emph{Proceedings of the Eleventh International Conference on
  Language Resources and Evaluation ({LREC} 2018)} (Miyazaki, Japan: European
  Language Resources Association (ELRA))
\bibAnnoteFile{uchida-etal-2018-cefr}

\bibitem[{\v{S}tajner(2021)}]{stajner-2021-automatic}
\v{S}tajner, S. (2021).
\newblock {Automatic Text Simplification for Social Good: Progress and
  Challenges}.
\newblock In \emph{Findings of the Association for Computational Linguistics:
  ACL-IJCNLP 2021} (Online: Association for Computational Linguistics),
  2637--2652.
\newblock \doi{10.18653/v1/2021.findings-acl.233}
\bibAnnoteFile{stajner-2021-automatic}

\bibitem[{W3C(2008)}]{W3C}
W3C (2008).
\newblock \emph{Web Content Accessibility Guidelines (WCAG) 2.0}
\bibAnnoteFile{W3C}

\bibitem[{Yimam et~al.(2018)Yimam, Biemann, Malmasi, Paetzold, Specia,
  {\v{S}}tajner et~al.}]{yimam-etal-2018-report}
Yimam, S.~M., Biemann, C., Malmasi, S., Paetzold, G., Specia, L.,
  {\v{S}}tajner, S., et~al. (2018).
\newblock {A Report on the Complex Word Identification Shared Task 2018}.
\newblock In \emph{Proceedings of the Thirteenth Workshop on Innovative Use of
  {NLP} for Building Educational Applications} (New Orleans, Louisiana:
  Association for Computational Linguistics), 66--78.
\newblock \doi{10.18653/v1/W18-0507}
\bibAnnoteFile{yimam-etal-2018-report}

\bibitem[{Yimam et~al.(2017)Yimam, {\v{S}}tajner, Riedl, and
  Biemann}]{YimamEtAl-2017-IJCNLP}
Yimam, S.~M., {\v{S}}tajner, S., Riedl, M., and Biemann, C. (2017).
\newblock {CWIG3G2 - Complex Word Identification Task across Three Text Genres
  and Two User Groups}.
\newblock In \emph{Proceedings of the Eighth International Joint Conference on
  Natural Language Processing (Volume 2: Short Papers)} (Asian Federation of
  Natural Language Processing), 401--407
\bibAnnoteFile{YimamEtAl-2017-IJCNLP}

\end{thebibliography}

%TC:endignore

%TC:ignore
\clearpage
\appendix
\section{Appendix I: Instructions for Annotators}

Below are {\bf N} sentences in English/Spanish/Portuguese, in each sentence there is a word marked in bold. Your task is to
write, in the space below each sentence,  \underline{single word} that has the same meaning as the one marked,
but is easier to understand. For example, in the sentence "At the same time, the rate of decline against
the dollar was {\bf attenuated}" the word {\bf attenuated} could be replaced by the easier-to-understand word
{\em decreased}. Write the replacement so that the replacement is valid in the given context. In our example,
{\em decreased} is correct while decrease would not be correct. In that case that it is not possible to replace
with a single word, then you can use a more complex substitution. For example in the sentence "The
dresses were {\bf Iranian}", the word {\bf Iranian} could be replaced by “{\bf from Iran}”. Replacements that involve a
gender change with respect to the marked word are also allowed in Spanish and Portuguese (Note that this is not applicable in English). 

\vspace{0.4cm}
Note 1: If you cannot find a simpler word then you must write the same
complex word in the answer area.

\vspace{0.4cm}
Note 2: You are allowed to use all kinds of lexical reference resources such as
dictionaries, thesaurus, etc., whether books or online, to do the task.

\vspace{0.4cm}
WARNING: In this task it is important that you follow the instructions to receive your payment.
By completing the task and clicking the purple button "Send" you affirm that you have read and agree to
the conditions of the information and consent form.

\vspace{0.4cm}
{\bf Information and Consent Form}

The study aims to collect examples of lexical simplification for English/Spanish/Portuguese. The data collected will be used
for research purposes only. You will read sentences in which a word considered complex will appear that
you should simplify by proposing another word that has the same meaning but is easier to understand.
The data collected will be used in a research project and will be provided to researchers who need it.
The results of this research may be published in scientific journals or conferences and may be used in
subsequent studies.
To participate in this experiment you should:
\begin{itemize}
\item a) Be a native English/Spanish/Portuguese speaker, 
\item b) Be at least 18 years old and competent to give consent.
\item c) Have read and understood this Information Form that explains the research project, 
\item d) You agree that the data collected will be used anonymously in the future, 
\item e) Agree to participate in the research described above.
\end{itemize}
Thanks for participating!

\clearpage

\section{Appendix II: Example Outputs in Each Language}

\begin{table}[h!]
    \begin{center}
    \scalebox{1.0}{
    \begin{tabular}{p{17cm}}
    \toprule
    Example 1
    \\ \midrule
Context and target: {\it A local witness said a separate group of attackers} {\bf disguised} {\it in burgas -- the head-to-toe robes worn by conservative Afghan women -- then tried to storm the compound.}\\
\\
Gold: \underline{concealed}, \underline{dressed}, hidden, camouflaged, changed, covered, masked, unrecognizable, converted, impersonated\\ 
\\
LSBert: {\bf dressed}, {\bf hidden}, hiding, {\bf covered}, buried\\
\\
TUNER: disguised, {\bf masked}\\
\\ \midrule
Example 2
\\ \midrule
Context and target: {\it War} {\bf maniacs} {\it of the South Korean puppet military made another grave provocation to the DPRK in the central western sector of the front on Thursday afternoon.}\\
\\
Gold: \underline{fanatics}, crazies, freaks, addicts, fans, lunatics, enthusiasts, fiends, fools, hawks, mongers, nuts, psychos\\ 
\\
LSBert: criminals, victims, machines, {\bf freaks}, people\\
\\
TUNER: {\bf lunatics}, madmans,  maniacs\\
\\ \midrule
Example 3
\\ \midrule
Context and target: {\it That prompted the military to} {\bf deploy} {\it its largest warship, the BRP Gregorio del Pilar, which was recently acquired from the United States.}\\
\\
Gold: \underline{send}, post, use, position,  employ, extend, launch, let loose, organize, redistribute, release, send out, set up, situate, station\\ 
\\
LSBert: {\bf launch}, dispatch, {\bf use}, develop, {\bf employ}\\
\\
TUNER: deploy\\\hline   
\end{tabular}
}
    \caption{Instances from the English part of the dataset together with the output of TUNER and LSBert systems for English. The suggested replacements in the gold data are presented in descending order with respect to the number of people who suggested them. The most frequently suggested ones are underlined (ties are possible). Suggestions by the systems that are present in the gold data are shown in bold.}
    \label{tab:OutputEN}
\end{center}    
\end{table}

\begin{table}[]
    \begin{center}
    \scalebox{1.0}{
    \begin{tabular}{p{17cm}}
    \toprule
    Example 1
    \\ \midrule
Context and target: {\it Las bancas (exedras) y balaustradas fueron hechas de m\'{a}rmol y el piso de granito, tambi\'{e}n hab\'{i}an cuatro fuentes de agua, faroles de bronce y jardines de flores que encantaban mucho a los} {\bf transe\'{u}ntes}\\
\\
Gold: \underline{peatones}, caminantes, paseantes, pasantes, personas que caminaban por ah\'{i}, viandantes, pasajeros, pedestres, personas con barandal, viajantes, las personas que pasaban \\

\\
LSBert: turistas, vecinos, {\bf pasajeros}, ciudadanos, visitantes\\
\\
TUNER: transe\'{u}ntes\\
\\ \midrule
Example 2
\\ \midrule
Context and target: {\it Conforme avanzaba el debate en el Congreso de Filadelfia, Lee iba asumiento una posici\'{o}n m\'{a}s favorable a la independencia total y no s\'{o}lo a la autonom\'{i}a del Imperio Brit\'{a}nico, su} {\bf convicci\'{o}n} {\it de la necesidad de la independencia logr\'{o} convencer a delegados de otras colonias e incluso persuadi\'{o} a sus propios electores de Virginia, temerosos que Lee pudiera llegar demasiado lejos.}\\
\\
Gold:  \underline{creencia}, \underline{seguridad}, \underline{certeza}, convencimiento, ideal, f\'{e}, persuaci\'{o}n, fuerte creencia, idea\\
\\
LSBert: {\bf idea}, {\bf creencia}, sentido, sentimiento, experiencia\\
\\
TUNER: {\bf certeza}, convicci\'{o}n\\
\\ \midrule
Example 3
\\ \midrule
Context and target: {\it La Abad\'{i}a de Fontenay es una abad\'{i}a francesa de Marmagne (departamento de C\^{o}te-d'Or), uno de los} {\bf monasterios} {\it m\'{a}s emblem\'{a}ticos de toda la arquitectura cisterciense.}\\
\\
Gold: \underline{conventos}, templos, las iglesias, ecierro religioso, claustros, lugares\\
\\ 
LSBert: {\bf lugares}, edificios, sitios, puntos, elementos\\
\\
TUNER: monasterios, {\bf conventos}\\
\hline      
    \end{tabular}
    }
    \caption{Instances from the Spanish part of the dataset together with the output of TUNER and LSBert systems for Spanish. The suggested replacements in the gold data are presented in descending order with respect to the number of people who suggested them. The most frequently suggested ones are underlined (ties are possible). Suggestions by the systems that are present in the gold data are shown in bold.}
    \label{tab:OutputES}
\end{center}    
\end{table}

\begin{table}[]
    \begin{center}
    \scalebox{1.0}{
    \begin{tabular}{p{17cm}}
    \toprule
    Example 1
    \\ \midrule
Context and target: {\it Quem n\~{a}o conseguir} {\bf esgotar} {\it o armazenamento de diesel puro n\~{a}o pode misturar com o b2 porque o produto ficaria fora de especifica\c{c}\~{a}o.}\\
\\
Gold: \underline{acabar}, esvaziar, acabar com, gastar, consumir, diminuir, zerar\\ 
\\
LSBert: recuperar, terminar, {\bf acabar}, completar, fazer\\
\\
TUNER: {\bf consumir}, esgotar\\
\\ \midrule
Example 2
\\ \midrule
Context and target: {\it Naquele pa\'{i}s a ave \'{e} considerada uma} {\bf praga}.\\
\\
Gold: \underline{peste}, epidemia, maldi\c{c}\~{a}o, doen\c{c}a, desgra\c{c}a, trag\'{e}dia, infesta\c{c}\~{a}o\\
\\
LSBert: amea\c{c}a, {\bf doen\c{c}a}, droga, delas, esp\'{e}cie\\
\\
TUNER: praga, {\bf peste}\\
\\ \midrule
Example 3
\\ \midrule
Context and target: {\it Autores de furto estariam} {\bf migrando} {\it para o roubo.}\\
\\
Gold: \underline{mudando}, passando, deslocando-se, indo, se direcionando, deslocando, se mudando\\
\\
LSBert: {\bf indo}, trabalhando, saindo, entrando, voltando\\
\\
TUNER: migrando, emigrando\\
\hline      
    \end{tabular}
    }
    \caption{Instances from the Portuguese part of the dataset together with the output of TUNER and LSBert systems for Portuguese. The suggested replacements in the gold data are presented in descending order with respect to the number of people who suggested them. The most frequently suggested ones are underlined (ties are possible). Suggestions by the systems that are present in the gold data are shown in bold.}
    \label{tab:OutputPT}
\end{center}    
\end{table}

%TC:endignore

\end{document}